\documentclass{article}

\usepackage{times}
\usepackage{helvet}
\usepackage{courier}
\usepackage{pdfsync}

\usepackage{graphicx} 
\usepackage{subfigure}
\usepackage{sidecap}
\usepackage{enumitem}

\usepackage{amsmath, amssymb}
\usepackage{amsthm}
\usepackage{natbib}

\usepackage{algorithm}
\usepackage{algpseudocode}
\usepackage[accepted]{icml2018}

\newtheorem{theorem}{Theorem}

\newtheorem{proposition}[theorem]{Proposition}
\theoremstyle{remark}

\newcounter{assumption}
\renewcommand{\theassumption}{A\arabic{assumption}}

\newenvironment{assumption}[1][]{\begin{trivlist}\item[] \refstepcounter{assumption}%
 \textbf{Assumption\ \theassumption\ #1} }{
 \ifvmode\smallskip\fi\end{trivlist}}

\renewcommand{\AA}{{\mathcal{A}}}

\newcommand{\XX}{{\mathcal{X}}}

\newcommand{\ZZ}{{\mathcal{Z}}}

\newcommand{\beq}{\begin{equation}}
\newcommand{\eeq}{\end{equation}}
\newcommand{\beqa}{\begin{eqnarray}}
\newcommand{\eeqa}{\end{eqnarray}}
\newcommand{\beqan}{\begin{eqnarray*}}
\newcommand{\eeqan}{\end{eqnarray*}}
\newcommand{\ben}{\begin{eqnarray*}}
\newcommand{\een}{\end{eqnarray*}}

\newcommand{\norm}[1]{\left\Vert#1\right\Vert}
\newcommand{\smallnorm}[1]{\Vert#1\Vert}

\newcommand{\Real}{\mathbb R}

\newcommand{\EE}[1]{{\mathbb E}\left[#1\right]}

\newcommand{\eps}{\varepsilon}

\newcommand{\ra}{\rightarrow}

\newcommand{\FF}{{\mathcal{F}}}

\newcommand{\cN}{{\mathcal{N}}}

\newcommand{\eqdef}{\triangleq}

\newcommand{\MM}{\mathcal{M}}

\newcommand{\cset}[2]{\left\{\,#1\,:\,#2\,\right\}}

\newcommand{\States}{\mathcal{X}}
\newcommand{\Actions}{\mathcal{A}}

\newcommand{\PKernel}{\mathcal{P}}
\newcommand{\RKernel}{\mathcal{R}}
\newcommand{\state}{x}
\newcommand{\action}{a}

\newcommand{\reward}{r}

\newcommand{\One}[1]{{\mathbb I}{\{#1\}}}

\newcommand{\actorparam}{{\theta^\mu}}
\newcommand{\criticparam}{{\theta^Q}}
\newcommand{\CKernel}{\mathcal{C}}

\newcommand{\stateind}[3]{\state^{(#2,#3)}_{#1}}
\newcommand{\numactions}{k}

\newcommand{\actionvalues}{u}

\newcommand{\figwidthtwo}{0.48\textwidth}
\newcommand{\figwidththree}{0.325\textwidth}

\newcommand{\todo}[1]{} 

\newif\ifSupp
\Supptrue

\icmltitlerunning{RL with Function-Valued Action Space for PDE Control}

\begin{document}

\twocolumn[
\icmltitle{Reinforcement Learning with Function-Valued Action Spaces \\ for Partial Differential Equation Control}

\begin{icmlauthorlist}
\icmlauthor{Yangchen Pan}{uofa,merl}
\icmlauthor{Amir-massoud Farahmand}{vector,merl}
\icmlauthor{Martha White}{uofa}
\icmlauthor{Saleh Nabi}{merl}
\icmlauthor{Piyush Grover}{merl}
\icmlauthor{Daniel Nikovski}{merl}
\end{icmlauthorlist}
\icmlaffiliation{uofa}{Department of Computing Science, University of Alberta, Edmonton, Canada}
\icmlaffiliation{merl}{Mitsubishi Electric Research Laboratories (MERL), Cambridge, USA}
\icmlaffiliation{vector}{Vector Institute, Toronto, Canada}
\icmlcorrespondingauthor{Yangchen Pan}{pan6@ualberta.ca}

\icmlkeywords{reinforcement learning, partial differential equation, PDE, high-dimensional}

\vskip 0.3in
]

\printAffiliationsAndNotice{}

\begin{abstract}
Recent work has shown that reinforcement learning (RL) is a promising approach to control dynamical systems described by partial differential equations (PDE). This paper shows how to use RL to tackle more general PDE control problems that have continuous high-dimensional action spaces with spatial relationship among action dimensions.
In particular, we propose the concept of \emph{action descriptors}, which encode regularities among spatially-extended action dimensions and enable the agent to control high-dimensional action PDEs. We provide theoretical evidence suggesting that this approach can be more sample efficient compared to a conventional approach that treats each action dimension separately and does not explicitly exploit the spatial regularity of the action space.
The action descriptor approach is then used within the deep deterministic policy gradient algorithm. 
Experiments on two PDE control problems, with up to $256$-dimensional continuous actions, show the advantage of the proposed approach over the conventional one.
\end{abstract}

\section{Introduction}
\label{sec:HDARL-Introduction}

This paper develops an algorithmic framework for handling reinforcement learning (RL) problems with high-dimensional action spaces.
We are particularly interested in problems where the dimension of the action space is very large or even infinite.
These types of problems naturally appear in the control of Partial Differential/Difference Equations (PDE), which have attracted attention because of their potential applications spreading over physical dynamic system \citep{jacques1971} and engineering problems, including the design of air conditioning systems \citep{Popescu2008aircondition}, modelling of flexible artificial muscles with many degrees of freedom \citep{kim2013bio}, traffic control~\citep{Richards1956,lit1955}, and the modelling of information flow in social networks \citep{haiyan2013socialpde}.

Many dynamical systems can be described by a set of Ordinary Differential Equations (ODE). Some examples are the dynamics describing inverted pendulum, robotic arm manipulator (with inflexible joints), and electrical circuits (in low-frequency regime for which the electromagnetic radiation is negligible).
The property of these systems is that their state can be described by a finite dimensional variable. The use of RL to control ODE-based problems, with either discretized action or continuous actions, has been widely presented in various RL works \citep{Sutton98,kober2013,DeisenrothNeumannPeters2013}. The success is most significant for problems where the traditional control engineering approaches may not fare well, due to the complexity of the dynamics, either in the form of nonlinearity or uncertainty.
There are, however, many other physical phenomena that cannot be well-described by an ODE, but can be described by a PDE. Examples are the distribution of heat in an object as a function of time and location, motion of fluids, and electromagnetic radiation, which are described by the heat, Navier-Stokes, and Maxwell's equations, respectively. The control of PDEs using data-driven approaches, including RL-based formulation, has just recently attracted attention and is a relatively an unexplored area~\citep{FarahmandNabiGroverNikovski2016,FarahmandNabiNikovski2017,BellettiHaziza2018,DuriezBruntonNoack2016}.

Control of PDEs has been investigated in conventional control engineering~\citep{KrsticSmyshlyaev2008,AhujaSuranaCliff2011, BorggaardBurnsSuranaZietsman2009,BurnsHeHu2016, BurnsHu2013,BruntonNoack2015}.
Despite the mathematical elegance of conventional approaches, they have some drawbacks that motivate investigating learning-based methods, in particular RL-based approaches.
Many conventional approaches require the knowledge of the PDE model, which might be difficult to obtain. Designing a controller for the model also requires a control engineer with an expertise in modeling and control of PDEs, which makes it even more challenging.
Furthermore, a hand-designed controller is brittle to changes in the geometry and parameters of the PDE, requiring repeated redesign of the controller to maintain  performance. This is impractical in many industrial applications, such as an air conditioning system that is deployed to a customer's home.
Moreover, many of the controller design tools are based on a linear PDE assumption, which ignores potentially useful nonlinear phenomena inherent in PDEs, such as those in fluid dynamics problems~\citep{FouresCaulfieldSchmid2014}. Designing a controller based on this simplified model might lead to a suboptimal solution.
The optimal controller is also typically designed for the quadratic cost function, as opposed to a more general objective, such as a user's comfort level for an air conditioning system.

It is desirable to have a controller design procedure that has minimal assumptions, does not require the knowledge of the PDE, and is completely data-driven. 
%
This can be achieved by formulating the PDE control problem as an RL problem, as has recently been shown~\cite{FarahmandNabiGroverNikovski2016,FarahmandNabiNikovski2017}, or some other flexible data-driven approaches such as the genetic programming-based method of~\citet{DuriezBruntonNoack2016}.

The PDE control problem is challenging because the state of a PDE is in theory an infinite-dimensional vector and very high-dimensional in computer simulations.
Moreover, many PDE control problems have infinite-dimensional continuous action, i.e., function-valued action. For example, if the PDE is controlled through its boundary condition, the action space is a function space defined over the continuous boundary.
Even though one may spatially discretize the action space and treat it as a multi-dimensional action space, such a solution leads to a very high-dimensional continuous action space and does not explicitly incorporate the spatial regularity of the PDE's action space. 

Previous work has addressed the problem of high-dimensionality of the state space in the RL-based PDE control~\citep{FarahmandNabiGroverNikovski2016,FarahmandNabiNikovski2017}, but their solutions have been limited to PDEs with a finite number of actions. This work focuses on the infinite-dimensionality of the action space. 
We formulate the PDE control with RL within the Markov Decision Process (MDP) formalism with function-valued state and action spaces (Section~\ref{sec:HDARL-Formulation}).
We introduce \emph{action descriptors} as a generic method to model actions in PDE control problems (Section~\ref{sec:HDARL-ActionDescriptors}).
Action descriptors allow us to scale to arbitrarily high dimensional continuous action spaces and capture the spatial regularities among action dimensions.
By benefiting from action descriptors, we propose a neural network architecture that is not changed with the increase of action dimensions; rather, it simply queries from a set of action descriptors (Section~\ref{sec:HDARL-FunctionalPG}). This is in contrast with conventional RL algorithms that directly output a high-dimensional action \emph{vector}. 
We provide some theoretical insights on why the proposed approach might have a better sample complexity through a covering number argument (Section~\ref{sec:HDARL-Theory}). 
Finally, we empirically verify the effectiveness of our architecture on two PDE control domains with up to $256$ dimensional continuous actions (Section~\ref{sec:HDARL-Experiments}).

\section{Reformulating PDE Control as an MDP}
\label{sec:HDARL-Formulation}



In this section, we first provide a PDE control example and briefly discuss how this can be viewed as an MDP. We also highlight the need to exploit the spatial regularities in the action space of PDEs. For more discussion, refer to~\citet{FarahmandNabiGroverNikovski2016}.

\subsection{Heat Invader: A PDE Example}

PDE control is the problem of modifying the behaviour of a dynamical system that is described by a set of PDEs.
An example of a PDE is the convection-diffusion equation, which describes the changes in the temperature in an environment, among several other other physical phenomena. 
Given a domain $\ZZ \subset \Real^d$ and time $t$, 
the temperature at location $z \in \ZZ$ at time $t$ is denoted by the scalar field $T(z,t)$. The convection-diffusion equation describes the evolution of the temperature as follows:
\begin{equation}
\label{eq:conv_diff_pde}
	\frac{\partial T}{\partial t}  = \nabla \cdot \frac{1}{P_e} \nabla T - \nabla \cdot (v T) + S.
\end{equation}
Here $S = S(z, t)$ is the heat source/sink, $v = v(z, t)$ is the velocity field describing the airflow in the domain,
and $\frac{1}{P_e}$ is a diffusivity constant. 
The gradient operator is $\nabla$, and the divergence operator is $\nabla \cdot$, which measures the outflow of a vector field. These operators are with respect to (w.r.t.) $z$.
We can further decompose the source term $S(z, t) = S(z, t;a) = S^o(z, t) + a(z ,t)$, where $S^o(z, t)$ is a source term that cannot be controlled (e.g., a disturbance) and $a(z ,t)$ is the action under our control.
The Heat Invader problem is a particular example of this convection-diffusion PDE over a 2D domain $\ZZ$ with a time-varying source $S^o(z,t)$~\citep{FarahmandNabiGroverNikovski2016}.

%

Since the location space $\ZZ$ is a continuous domain, the action $a(\cdot,t)$ is a function lying in some function space. The dimension of this action space, however, depends on how we actuate the PDE. As a concrete example, if we can control the temperature of the airflow going through an inlet in an air conditioning system, the action is the scalar temperature of the inflow air and the action space is one-dimensional real-valued vector space (assuming we can control the temperature with arbitrary precision).
But we may also be able to finely control the temperature of walls of a room by a distributed set of minuscule heaters/coolers within the wall (this technology might not exist today, but is perceivable). In this case, the action is the temperature of each tiny heater/cooler in the wall, so the action space would be an extremely high dimensional vector space.

Generally, one may control a PDE by setting its boundary condition or by defining a source/sink within its domain. 
The boundary of a PDE has a dimension one smaller than the domain of a PDE (under certain regularities of the geometry of the boundary). So for a 3D PDE (e.g., a room and its temperature field), the boundary is a 2D domain (e.g., the walls). 
An action defined over the boundary, without further constraint, is a function with a domain that is uncountably large. A similar observation holds when we control a source/sink within the domain.

\subsection{PDE Control as an MDP}
To relate the PDE control problem to the reinforcement learning setting, we first provide a brief overview of MDP~\citep{Sutton98,SzepesvariBook10,Bertsekas2013}.
%
An MDP consists of $(\States, \Actions, \PKernel, \RKernel, \gamma)$, where $\States$ denotes the state space, $\Actions$ denotes the action space, $\PKernel: \States \times \Actions \times \States \rightarrow \MM(\XX)$ is the transition probability kernel (with $\MM(\XX)$ being the space of probability distributions defined over $\XX$), $\RKernel: \XX \times \AA \times \XX \ra \MM(\Real)$ is the reward distribution, and $\gamma \in [0, 1)$ is the discount factor. At each discrete time step $t = 1,2,3,...$, the agent selects an action according to some policy $\pi$ and the environment responds by transitioning into a new state $\state_{t+1}$ sampled from $\PKernel (\cdot | \state_t, \action_t)$, and the agent receives a scalar reward $r_{t+1}$ samples from $ \RKernel(\cdot | \state_t, \action_t, \state_{t+1})$. The agent's goal is to maximize discounted cumulative sum of rewards from the current state $\state_t$. Typically in the RL literature, both the state space and action spaces are finite dimensional vector spaces, for example both are subsets of an Euclidean space, $\Real^p$ and $\Real^k$, respectively. 
The state of a PDE, however, is an infinite-dimensional vector space. Thus to describe such problems, we need to work the MDPs with infinite dimensional state and action spaces.\footnote{The discounted MDP framework works fine with general state spaces, under certain measurability conditions, cf., Sections 5.3 and 5.4 and Appendix C of~\citet{Bertsekas2013}. The conditions would be satisfied for bounded Borel measurable reward function and the Borel measurable stochastic kernel.}


Consider now the PDE control problem, and how it can be formalized as an agent finding an optimal policy for an MDP $(\States, \Actions, \PKernel, \RKernel, \gamma)$. 
In~\eqref{eq:conv_diff_pde}, the state and action are infinite dimensional vectors (functions) $\state_t = T(\cdot, t)$ and $\action_t = a(\cdot, t)$. 
For example in the Heat Invader problem, the state is the current temperature at all locations, and the action changes the temperature at each location.
Both state and action are functions belonging to some function space defined over the domain $\ZZ$ of PDE, e.g., the space of continuous function $\mathcal{C}(\ZZ)$, the Sobolev space, etc.
The particular choice of function space depends on the PDE and its regularity; we denote the space by $\FF(\ZZ)$. 

 

The dynamics of the MDP is a function of the dynamics of PDE. One difference between the MDP framework and PDE is that the former describes a discrete-time dynamical system whereas the latter describes a continuous-time one. One may, however, integrate the PDE over some arbitrary chosen time step $\Delta_t$ to obtain a corresponding partial \emph{difference} equation $x_{t+1} = f(x_t, a_t)$ for some function $f: \FF(\ZZ) \times \FF(\ZZ) \ra \FF(\ZZ)$, which depends on the PDE.
For the convection-diffusion PDE~\eqref{eq:conv_diff_pde} with the state being denoted by $x \in \FF(\ZZ)$ (instead of temperature $T$), we have
$
	f(x,a)  = \int_{t=0}^{\Delta_t} \nabla \cdot \frac{1}{P_e} \tilde{x} - \nabla \cdot (v \tilde{x}) + S(z, t;a) \mathrm{d}t 
$
with the initial state of $\tilde{x}$ at $t = 0$ being $x$ and the choice of $S(z,t;a)$ depending on $a$.\footnote{Note that this step requires the technical requirement of the existence of the solution of a PDE, which has not been proven for all PDEs, e.g., the Navier-Stokes equation. Also we assume that the action remains the same for that time period.}
So we might write $\PKernel (\state | \state_t, \action_t) = \delta (\state - f(\state_t, \action_t))$, where $\delta(\cdot)$ is the Dirac delta function.
More generally, if there is stochasticity in the dynamics, for example if the constants describing the PDE are random, the temporal evolution of the PDE can be described by a transition probability kernel, i.e.,
$x_{t+1} \sim \PKernel(\cdot|x_t, a_t)$.

The remaining specification of the MDP, including the reward and the discount factor, is straightforward. For example, the reward function in the Heat Invader problem is designed based on the desire to keep the room temperature at a comfortable level while saving energy:
\begin{align}\label{heatrewardfunc}
r(\state_t, \action_t, &\state_{t+1}) =  - \text{cost}(\action_t)\\
& - \int_{z \in \ZZ}^{} \mathbb{I} ({|T(z, t+1)| > T^\ast(z, t+1)}) \mathrm{d}z, \nonumber
\end{align}
where $\mathbb{I}(\cdot)$ is an indicator function, $T^\ast(\cdot)$ is a predefined function describing the acceptable threshold of deviation from a comfortable temperature (assumed to be $0$), and $\text{cost}(\cdot)$ is a penalty for high-cost actions.
\ifSupp
Refer to Appendix~\ref{sec:heat_details} for more detail.
\else
Refer to the supplementary material for more detail.
\fi

Reinforcement learning algorithms, however, are not designed to learn with infinite-dimensional states and actions. 
We can overcome this problem by exploiting the spatial regularities in the problem, and beyond PDEs, exploiting general regularities between the dimensions of the states as well as actions. We provide more intuition for these regularities before introducing MDPs with action-descriptors, a general subclass of infinite-dimensional MDPs that provides a feasible approach to solving these infinite-dimensional problems.


\subsection{Exploiting Spatial Regularities in PDEs}

%
One type of regularity particular to PDEs is the spatial regularity of their state.
This regularity becomes apparent by noticing that the solution of a PDE, which is typically a 1D/2D/3D scalar or vector field, is similar to a 1D/2D/3D image in a computer vision problem.
%
This similarity has motivated some previous work to design RL algorithms that directly work with the PDE's infinite dimensional state vector, or more accurately its very high-dimensional representation on a computer, by treating the state as an image~\citep{FarahmandNabiGroverNikovski2016,FarahmandNabiNikovski2017}.
\citet{FarahmandNabiGroverNikovski2016} suggest using the Regularized Fitted Q-Iteration (RFQI) algorithm~\cite{FarahmandACC09} with a reproducing kernel Hilbert space (RKHS) as a value function approximator. For that approach, one only needs to define a kernel between two image-like objects.
\citet{FarahmandNabiNikovski2017} suggest using a deep convolution network (ConvNet) as the estimator of the value function. ConvNets are suitable to exploit spatial regularities of the input and can learn domain-specific features from image-like inputs.

Even though these previous approaches can handle high-dimensional states in PDE control problems, they are limited to a finite number of actions. Their approach does not exploit the possible regularities in the \emph{action space} of a PDE. For example, some small local changes in a controllable boundary condition may not change the solution of the PDE very much. In that case, it makes sense to ensure that the actions of nearby points on the boundary be similar to each other.
The discretization-based approach~\citep{FarahmandNabiGroverNikovski2016,FarahmandNabiNikovski2017} ignores the possibility of having a spatial regularity in the action space.
We next describe how we can exploit these regularities---as well as make it more feasible to apply reinforcement learning algorithms to these extraordinarily high-dimensional continuous action problems---by introducing the idea of action descriptors. 


%


\section{MDPs with Action Descriptors}
\label{sec:HDARL-ActionDescriptors}

We consider an MDP formulation where the state space $\States$ and action space $\Actions$ can be infinite dimensional vector spaces, e.g., the space of continuous functions over $\ZZ$. 
%
Procedurally, the agent-environment interaction is as usual: at each step, the agent selects a function $\action_t \in \Actions$ at the current state $\state_t \in \States$, traverse to a new state $\state_{t+1}$ according to the dynamics of the PDE, and receives reward $r_{t+1}$.




Even though the infinite dimensional state/action space MDPs provide a suitable mathematical framework to talk about control of PDEs, a practically implementable agent may not be able to provide an infinite dimensional action as its output, i.e., providing a value for all uncountably infinite number of points over $\ZZ$. 
Rather, it may only select the values of actions at a finite number of locations in $\ZZ$, and convert those values to an infinite dimensional action appropriate as the control input to the PDE.
Consider the Heat Invader problem. There might be a fine, but finite, grid of heater/cooler elements on the wall whose temperature can be separately controlled by the agent.  
Each of the elements is spatially extended (i.e., each element covers a subset of $\ZZ$), and together they define a scalar field that is controlled by the agent. The result is an infinite dimensional action, an appropriate control input for a PDE, even though the agent only controls a finite, but possibly very large, number of values.
In some cases, the set of controllable locations are not necessarily fixed, and might change. For example, if an air conditioner is moved to some other location which has never placed before, the agent should ideally still be able to control them.

We propose to model this selection of action-dimensions (or the location of where the agent can exert control) using \emph{action descriptors}. For the Heat Invader problem, the action descriptors correspond to the spatial locations $z$ of the air conditioners; more generally, they can be any vector describing an action-dimension in the infinite-dimensional action space. The action descriptors help capture regularities across action dimensions, based on similarities between these vectors.

To be concrete, let $\ZZ$ be the set of locations in the domain of PDE, e.g., $\ZZ = [0,1]^2$ for the 2D convection-diffusion equation~\eqref{eq:conv_diff_pde}.
An action descriptor is $c \in \ZZ$ and determines the location where the action can be selected by the agent. We use the more general term action descriptor, rather than actions locations, as in other PDEs, $\ZZ$ may represent other regularities between actions. The set of all action descriptors is the finite and ordered set $\mathcal{C} = (c_1, \dotsc, c_\numactions )$, for finite $\numactions$.
Given each action descriptor $c_i$ and the state $x$, the agent's policy $\pi: \mathcal{X} \times \ZZ \ra \Real$ (in case of deterministic policies) generates an action scalar $u^{(i)} \in \Real$, i.e, $u^{(i)} = \pi(x, c_i)$.
Recalling that the action of a PDE belongs to $\FF(\ZZ)$, an infinite dimensional action space, we have to convert these action scalars to an action that can be fed to the underlying PDE.
We use an \emph{adapter} for this purpose.
An adapter is a mapping from the set of action scalars $\actionvalues = (u^{(1)}, \dotsc, u^{(\numactions)})$ and the action descriptors $\mathcal{C}$ to a function defined over $\ZZ$, i.e., $I: \mathcal{C} \times \Real^\numactions \ra \FF(\ZZ)$.
The infinite dimensional action given to the PDE is $a_t = I(\mathcal{C}, \actionvalues_t)$.

%

There are many ways to define an adapter, which can be thought of as a decoder from a finite-dimensional code to a function-valued code.
Linear interpolators are one particular class of adapters. Given a set of (fixed) weighting functions $w_i: \ZZ \ra \Real$, the linear interpolator is
\begin{align}
\label{sec:HDARL-Adapter-LinearInterpolator}
	I( \mathcal{C}, \actionvalues):
	z \mapsto 
	\sum_{i=1}^\numactions w_i(z) u^{(i)}.
\end{align}
For instance, we may define the weighting function as Gaussians with the centres on the action descriptors, i.e.,
$w_i(z) = \exp( -\frac{ \smallnorm{z - c_i}^2}{2 \sigma^2} )$.

Another choice, which we use in our experiments, is based on partitioning of the domain $\ZZ$ around the action descriptors.
Given the action descriptors $\mathcal{C}$, define a partition of $\ZZ$ and denote it by $(A_1, \dotsc, A_\numactions)$, i.e., $\bigcup A_i = \ZZ$ and $A_i \cap A_j = \emptyset$ for $i \neq j$.
For example, $A_i$ might be a rectangular-shaped region in $\ZZ$ and $c_i$ being its centre. Another example would be Voronoi diagram corresponding to centres in $\mathcal{C}$, which is commonly used in the finite element method.
Having this partitioning, we define the weighting functions as $w_i(z) = \One{z \in A_i}$.
With this choice of adapter,  the action given to the MDP is
\begin{align}
\label{sec:HDARL-Action-through-Adapter-Partitionbased}
	a_t =
	I( \mathcal{C}, \actionvalues) = 
	\sum_{i=1}^\numactions \One{z \in A_i} \pi(x_t, c_i).
\end{align}
To build physical intuition, consider the Heat Invader problem again. The action descriptors $\mathcal{C}$ correspond to the locations of the (centre of) heater/cooler elements of the air conditioners. Furthermore, a partition $A_i$ corresponds to the spatially-extended region where each element occupies. When we set the temperature at location $c_i$ to a certain value $u^{(i)}$ in~\eqref{sec:HDARL-Action-through-Adapter-Partitionbased}, the value of the whole region $A_i$ takes the same value $u^{(i)}$. 

The action descriptor formulation allows the agent to exploit the spatial regularity of the action.
For example if the variations of the action applied within a subregion $\ZZ_0 \subset \ZZ$ does not cause much change in the dynamics, it is sufficient to learn a $\pi(\cdot,c)$ that varies slowly within $\ZZ_0$.
The learning agent that has access to $\mathcal{C}$ and can learn a mapping $\pi(x,c_i)$, as in~\eqref{sec:HDARL-Adapter-LinearInterpolator} or \eqref{sec:HDARL-Action-through-Adapter-Partitionbased},
can potentially exploit this regularity.
This can be contrasted with a non-adaptive interpolation-only agent where the action is chosen as $a_t = \sum_{i=1}^\numactions \One{z \in A_i} \pi_i(x_t)$. In this scheme we choose the action at each partition without considering their relative locations through a series of separately learned $\pi_i(x)$. 
Such an agent cannot explicitly benefit from the spatial regularity of the action.
We develop this argument more rigorously in Section~\ref{sec:HDARL-Theory}.
Another benefit of the action description formulation is that as long as $I( \mathcal{C}, \actionvalues)$ can work with variable-sized sets $\mathcal{C}$ and $\actionvalues$, it allows variable number of spatial locations to be queried. This can be helpful when the physical actuators in the environment are added or removed.

This formulation is a strict generalization of more standard settings in reinforcement learning, such as a finite-dimensional action space $\AA = \Real^\numactions$. The domain for actions is $\ZZ = \{1, \dotsc, \numactions \}$ and we do not subselect locations, $\mathcal{C} = \ZZ$, requiring instead that all action dimensions are specified. The adapter simply returns action $\pi_i(x_t)$ for action index $i$, using $A_i = \{i\}$ and $a_t = \sum_{i=1}^\numactions \One{z \in A_i} \pi_i(x_t)$.
\section{PDE Control with RL Algorithms}
\label{sec:HDARL-FunctionalPG}

There has been some work addressing continuous action spaces, including both with action-value methods \citep{baird1993reinforcement,gaskett1999qlearning,delrmillan2002continuous,vanhasselt2007reinforcement} and policy-based methods \citep{schulman2015,WilliamLevine2016,silverlever2014,lillicrap2016,schulman2016high}. These methods, however, do not scale with extremely high-dimensional action spaces, because they either have to optimize the action-value function over a high-dimensional action space or output a high-dimensional action vector. These approaches also do not explicitly exploit regularities between actions. Some work for high-dimensional discrete (finite) action spaces do extract action embeddings  \citep{sunehag2015deep,he2015deep,dulac2015deep} or impose factorizations on the action space \citep{sallans2004reinforcement,dulac2012fast,pazis2011generalized}. These methods, however, are specific to large sets of discrete actions.
Existing methods cannot be directly applied to learning for these (extremely) high-dimensional continuous action problems with regularities. Next we discuss how to modify a policy gradient method to extend to this setting. 


For our MDP problem formulation with action descriptors, we propose to learn a policy function that receives the state along with action descriptors and outputs actions or probabilities over actions. Consider the policy parameterization $\pi_\actorparam : \States \times \ZZ \ra \mathbb{R}$.
For a given state $\state_t$, under the infinite-dimensional MDP formalism, the selected action $\action_t = \pi_\actorparam(\state_t, \cdot)$ which is function-valued. With the action descriptors set $\mathcal{C}$, the policy outputs the $i$th action component by evaluating $\pi_\actorparam(\state_t, c_i), c_i \in \mathcal{C}$ and hence we are able to get action scalars $\actionvalues_t \in \Real^{|\mathcal{C}|}$. 
Although a distribution over such functions could be maintained, for simplicity in this preliminary work, we focus on deterministic policies. 

The Deterministic Policy Gradient algorithm \cite{lillicrap2016} provides a method to learn such a policy.
For finite-dimensional action spaces $\Actions$, 
let $\pi_\actorparam(\cdot) : \States \rightarrow \Actions$ be the actor network parameterized by $\actorparam$, $Q(\cdot, \cdot; \criticparam) : \States \times \Actions \rightarrow \mathbb{R}$ be the critic network parameterized by $\criticparam$, and $d_{\pi_\actorparam}(\cdot)$ be corresponding stationary distribution. 
Similarly to the stochastic case \citep{SuttonMcAleesterSinghMansour2000}, the goal is to maximize the expected average reward, under that policy
\begin{equation*}
J(\pi_\actorparam) = \int_{\States \times \States} d_{\pi_\actorparam}(\state) r(\state, \pi_\actorparam(\state), \state')  \mathrm{d}\state \mathrm{d}\state' .
\end{equation*}
The Deterministic Policy Gradient theorem \citep[Theorem 1]{lillicrap2016,silverlever2014} shows that the gradient of this average reward objective, under certain conditions,
is approximately the expected value, across states, of $\nabla_\actorparam Q(\state,\pi_\actorparam(\state); \criticparam)$. Using the chain rule, this provides
a straightforward gradient ascent update for $\actorparam$: $\nabla_\action Q(\state,\action; \criticparam)|_{\action=\pi_\actorparam(\state)} \nabla_\actorparam \pi_\actorparam(\state)$, where $\nabla_a Q(\state,a; \criticparam)|_{a=\pi_\actorparam(\state)}$ is the Jacobian matrix generated by taking gradient of each action component with respect to the actor network parameters.

We can extend this algorithm to use action descriptors as shown in Algorithm~\ref{alg_ddpgpde}, which can efficiently scale to extremely high-dimensional continuous actions while capturing the intrinsic regularities. The change to DDPG is simply to modify the actor network, to input both states and action descriptors and output an action scalar. DDPG, on the other hand, would only receive the state and output a $k$-dimensional vector $u \in \Real^k$ to specify the action scalars. To get the $k$ action scalars with our actor network, the network is evaluated $k$ times with each action descriptor.

\begin{algorithm}[tb!]
\small
	\caption{DDPG with Action Descriptors}
	\label{alg_ddpgpde}
	\begin{algorithmic}
		\State Initialize a random process $\mathcal{N}$ for exploration\\
		Initialize an empty buffer $B$ for experience replay\\
		Initialize actor and critic networks (e.g., with Xavier initialization)
		$\pi(\cdot,\cdot; \actorparam): \States \times \ZZ \ra \Real, Q(\cdot,\cdot; \criticparam): \States \times \Real^k \ra \Real$ 
		and target actor and critic networks $\pi(\cdot,\cdot; \actorparam'), Q(\cdot,\cdot; \criticparam')$
		\State Define a set of action descriptors $\CKernel, |\CKernel| = k$ \\
		and target network update rate $\tau$
		\For {t = 1, 2, ...}
		\State Observe $\state_t$, compute action scalars $\actionvalues_t = [ \pi(\state_t,c_1; \actorparam), \dotsc, \pi(\state_t,c_k; \actorparam) ] + \mathcal{N} \in \Real^k$
		\State Execute action $a_t = I(\mathcal{C}, \actionvalues_t)$, and transition to state $\state_{t+1}$ and get reward $\reward_{t+1}$
	    \State Add sample $(\state_t, \actionvalues_t, \state_{t+1}, \reward_{t+1})$ to $B$
		\State $B_N \gets$ a mini-batch of $N$ samples from $B$
		\For {$(\state_i, \actionvalues_i, \state_{i+1}, \reward_{i+1}) \in B_N$}
		\State $u' = [\pi(\state_i,c_1; \actorparam'), \ldots, \pi(\state_i,c_k; \actorparam')]$
		\State set target $y_i = r_{i+1}+ \gamma Q(\state_{i+1}, u'; \criticparam')$ 
		\EndFor
		\State Update the critic by minimizing the loss: 
		\State $L = \tfrac{1}{N} \sum_{i=1}^{N} (y_i - Q(\state_i, \actionvalues_i; \criticparam))^2$ 
		\State Update the actor by gradient ascent, with gradient: 
		\State For $f(\actorparam) =  [\pi(\state_i,c_1; \actorparam), \ldots, \pi(\state_i,c_k; \actorparam)]$
		\State $\frac{1}{N} \sum_{i=1}^{N} \nabla_u Q(x,u; \criticparam)|_{u=f(\actorparam)} \nabla_\actorparam f(\actorparam)$ 
		\State Update target network parameters:
		\State $\actorparam' \gets (1-\tau) \actorparam' + \tau \actorparam$
		\State $\criticparam' \gets (1-\tau) \criticparam' + \tau \criticparam$
		\EndFor
	\end{algorithmic}
\end{algorithm}

\section{Theoretical Insights}
\label{sec:HDARL-Theory}

We provide theoretical evidence showing that learning a policy that explicitly incorporates the action location (or descriptor) $z$ might be beneficial compared to learning many separate policies for each action location.
The evidence for this intuitive result is based on comparing the covering number of two different policy spaces. 
The first is the space of policies with certain spatial regularity (Lipschitzness in $z$) explicitly encoded. The other is the policy space where the spatial regularity is not explicitly encoded. 
The covering number is a measure of complexity of a function space and appears in the estimation error terms of many error upper bounds, both in supervised learning problems~\citep{Gyorfi02,SteinwartChritmann2008} and in RL~\citep{AntosSzepesvariML08,LazaricGhavamzadehMunos2016,FarahmandGhavamzadehSzepesvariMannor2016}.
Note that the covering number-based argument is only one part of an error upper bound---even in the supervised learning theory \citep[page 61]{Mohri2012mlfund}. Due to complications arising from exploration strategy and convergence issues, to the best of our knowledge, there is no estimation error bound for deep reinforcement learning algorithms so far. Therefore, our results only provide a possible mathematical insight rather than a complete picture of the sample efficiency bound.



To formalize, consider a policy $\pi_\theta: \XX \times \ZZ \ra \Real$, parameterized by $\theta \in \Theta$. Let us denote this space by $\Pi_L$.  
We make the following assumptions regarding its regularities (refer to \ifSupp Appendix~\ref{sec:HDARL-Appendix-Theory} \else the supplementary material \fi for the definition of the covering number and the proof of the result).
\begin{assumption}
\label{ass:PiL-Regularities}
	 The following properties hold for the policy space $\Pi_L$:
	 	\vspace{-0.3cm}
	\begin{itemize}[leftmargin=*]
		\setlength\itemsep{0mm}
		\item For any fixed action location $z \in \ZZ$, the $\eps$-covering number of $\Pi_L \rvert_z \eqdef \cset{x \mapsto \pi_\theta(x,z)}{\theta \in \Theta}$  is $\cN(\eps)$. 
		\item The policy $\pi_\theta$ is $L$-Lipschitz in the action location $z$ uniformly in $\theta \in \Theta$ and $x \in \XX$, i.e., $|\pi_\theta(x,z_1) - \pi_\theta(x,z_2)| \leq L \norm{z_1 - z_2}$ for any $z_1, z_2 \in \ZZ$ and any $x \in \XX$. The domain $\ZZ$ is a bounded subset of $\Real^d$.
	\end{itemize}	
\vspace{-0.3cm}
\end{assumption}

%
%

We think of $\Pi_L$ as the policy space to which the optimal policy, or a good approximation thereof, belongs, but we do not know which member of it is the actual optimal policy. The role of any policy search algorithm, DDPG included, is to find that policy within $\Pi_L$. The stated assumptions on $\Pi_L$ describe certain types of regularities of $\Pi_L$, which manifest themselves both in the complexity of the policy space for a fixed location $z$ (through the covering number $\cN(\eps)$), and its Lipschitzness as the location parameter varies.
Note that we have not proved that the optimal policy for the Heat Invader problem, or any PDE control problem for that matter, in fact satisfies these regularities.

We would like to compare the $\eps$-covering number of $\Pi_L$ with the $\eps$-covering number of a policy space that does not explicitly benefit from the Lipschitzness of $\Pi_L$, but still can provide an $\eps$-approximation to any member of $\Pi_L$. 
This policy might be seen as the extreme example of the policy used by the conventional DDPG (or any other policy search algorithm) where each action dimension is represented separately. In other words, for having $N$-dimensional action, we have $N$ different function approximators.
Let us introduce some notations in order to define this policy space more precisely.

Consider a set of locations $\{c_i\}_{i=1}^{M_\eps}$ and their corresponding partition $\{A_i\}_{i=1}^{M_\eps}$ with resolution  $\frac{\eps}{2L}$. This means that for each $z \in \ZZ$, there exists a $c_i \in A_i$ such that the distance of $z$ to $c_i$ is less than $\frac{\eps}{2L}$ and $z \in \AA_i$.
The number of required partition is $M_\eps = c (\frac{2L}{\epsilon})^d $, for some constant $c > 0$, which depends on the choice of distance metric and the geometry of $\ZZ$ (but not $\eps$).
%
%
Define the following policy space (cf.~\eqref{sec:HDARL-Action-through-Adapter-Partitionbased}):
\begin{align*}
\underline{\Pi} = \Big \{ &\pi_{\underline{\theta}}(x,z) = \sum_{i=1}^{M_\eps} \pi_{\theta_i}(x, c_i) \One{ z \in A_i}:  \\
							  &\pi_{\theta_i} \in \Pi_L \rvert_{c_i}, i=1, \dotsc, M_\eps \Big\}.
\end{align*}
This is the policy space where each action location is modeled separately, and it is allowed to be as flexible as any policy in $\Pi_L$ with a fixed action location. But this policy space does not restrict the policy to be Lipschitz in $\ZZ$, so it is more complex than $\Pi_L$. 
The following proposition compares the complexity of $\underline{\Pi}$ and $\Pi_L$ in terms of the logarithm of their covering numbers (metric entropy).

\begin{proposition}
\label{prop:HDARL-CoveringNumber-LipschitzClasses}
	Consider two policy spaces $\Pi_L$ and $\underline{\Pi}$, as defined above.
	Suppose that $\Pi_L$ satisfies Assumption~\ref{ass:PiL-Regularities}.
	It holds that for any $\eps > 0$, the policy space $\underline{\Pi}$ provides an $\eps$-cover of $\Pi_L$.
	Furthermore, for some $c_1, c_2 > 0$, independent of $\eps$, the following upper bounds on the logarithm of the covering number hold:
	\vspace{-0.45cm}
	\begin{align*}
		& \log \cN(\eps,\Pi_L) \leq c_1 \left (\frac{L}{\eps} \right )^d + \log \cN(\eps), \\
		& \log \cN(\eps,\underline{\Pi}) \leq c_2 \left (\frac{L}{\eps} \right)^d \log \cN(\eps/2).
	\end{align*}
	\vspace{-0.6cm}
\end{proposition}
The covering number of $\underline{\Pi}$ grows faster than that of $\Pi_L$. This is intuitive as the former imposes less restriction on its members, i.e., no Lipschitzness over $\ZZ$.
To give a more tangible comparison between two results, suppose that $\log N(\eps) = c \eps^{-2\alpha}$ for some $c > 0$ and $0 \leq \alpha < 1$.\footnote{
This choice of metric entropy holds for some nonparametric function spaces, such as Sobolev spaces and some RKHS. We do not show that this covering number result actually holds for the policy space $\Pi_L \rvert_z$, so at this stage this is only an example.}
In that case, the metric entropy of $\Pi_L$ behaves as $\eps^{-\max\{d,2\alpha\}}$ whereas that of $\underline{\Pi}$ behaves as $\eps^{-(d+2 \alpha)}$.

Since there is no error bound for DDPG, we cannot compare the error bounds. But for some estimation problems such as regression or value function estimation with a function approximator with the metric entropy of $\log N(\eps) = c \eps^{-2\beta}$, the optimal error bound behaves as $O(n^{-\frac{1}{1+\beta}})$, with $n$ being the number of samples~\cite{YangBarron1999,FarahmandGhavamzadehSzepesvariMannor2016}.
Taking this as a rough estimate, the ratio of the error rates of $\Pi_L$ to that of $\underline{\Pi}$ is
\[
	n^\frac{- 2 \min\{2\alpha, d\} }{(2 + 2\alpha + d ) (2 + \max \{2\alpha, d \}) }
\]
which becomes significant as $\alpha$, the complexity of $\Pi_L \rvert_z$, grows.
For example, when $\alpha = 1$ and $d = 3$, the error rate of a method that uses $\Pi_L$ as the function approximator is $n^{1/6}$ faster than the other's.
This suggests the possible benefit of explicitly incorporating the spatial regularity in the $\ZZ$ space, in terms of sample complexity of learning.

\section{Experiments}
\label{sec:HDARL-Experiments}
We empirically show that our approach, which can exploit spatial regularity, can be easily scaled to large continuous action dimensions while maintaining competitive performance. We compare DDPG and DDPG with separate Neural Networks (NN) for each action component, to our DDPG with Action Descriptors on two domains: 
a simple PDE Model domain, and the Heat Invader problem as described throughout this work. The latter is a more difficult problem with more action regularities.

\subsection{Results on the PDE Model Domain}



\begin{figure*}[t]
	\subfigure[PDEModel: action dimension = 36]{
		\includegraphics[width=\figwidththree]{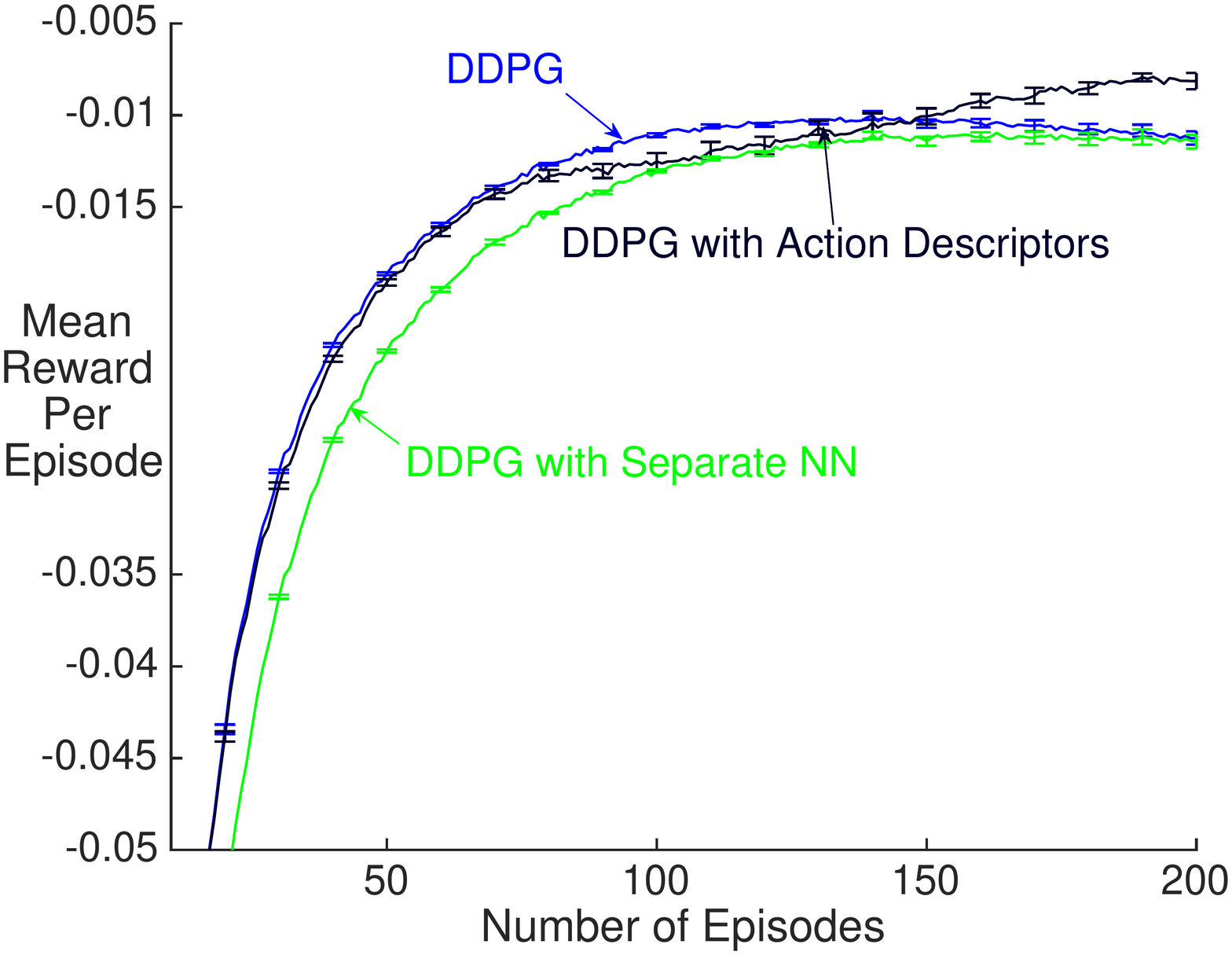}\label{fig:pdemodel_36d_l2}}
	\subfigure[PDEModel: action dimension = 100]{
		\includegraphics[width=\figwidththree]{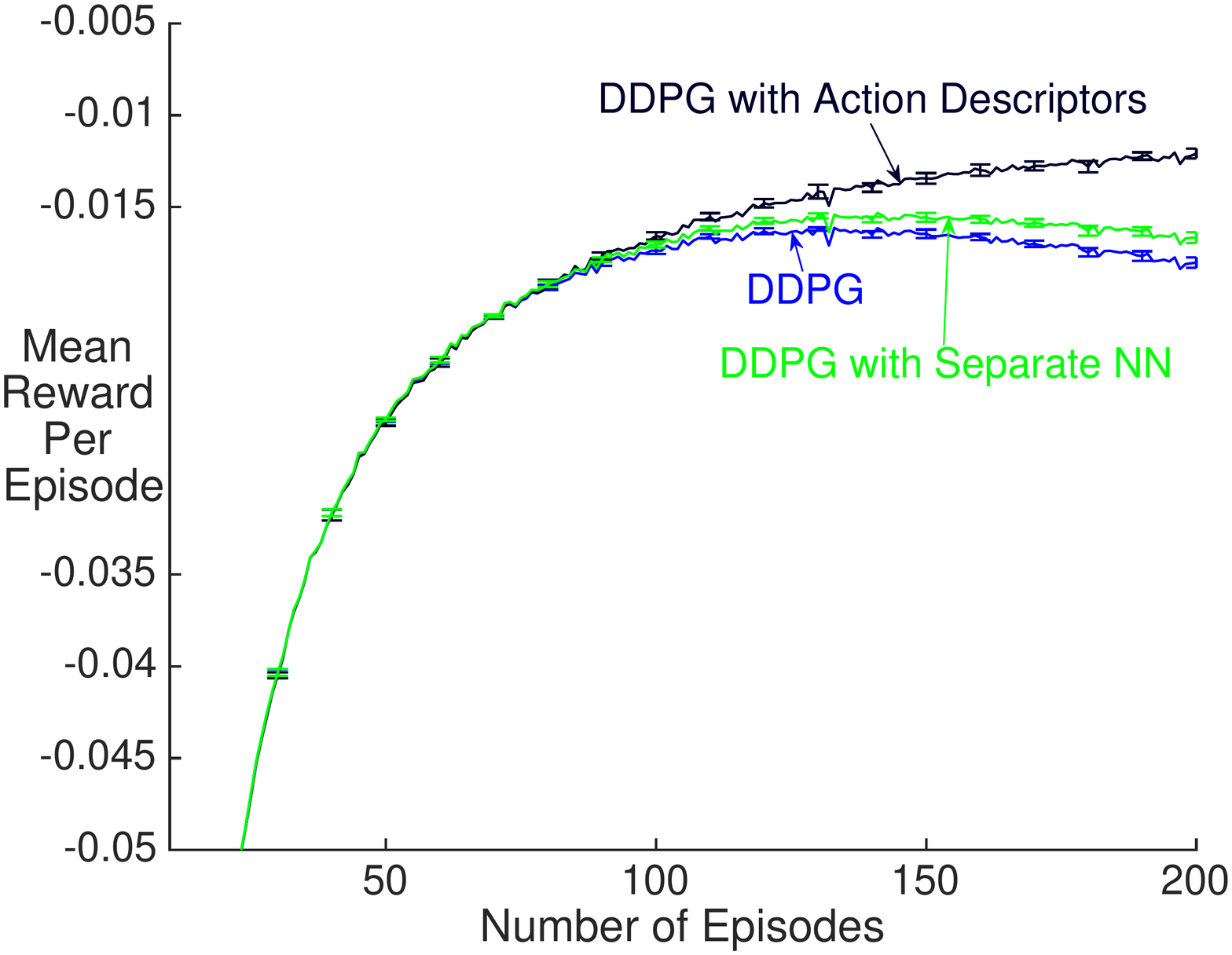}\label{fig:pdemodel_100d_l2}}
	\subfigure[PDEModel: action dimension = 256]{
		\includegraphics[width=\figwidththree]{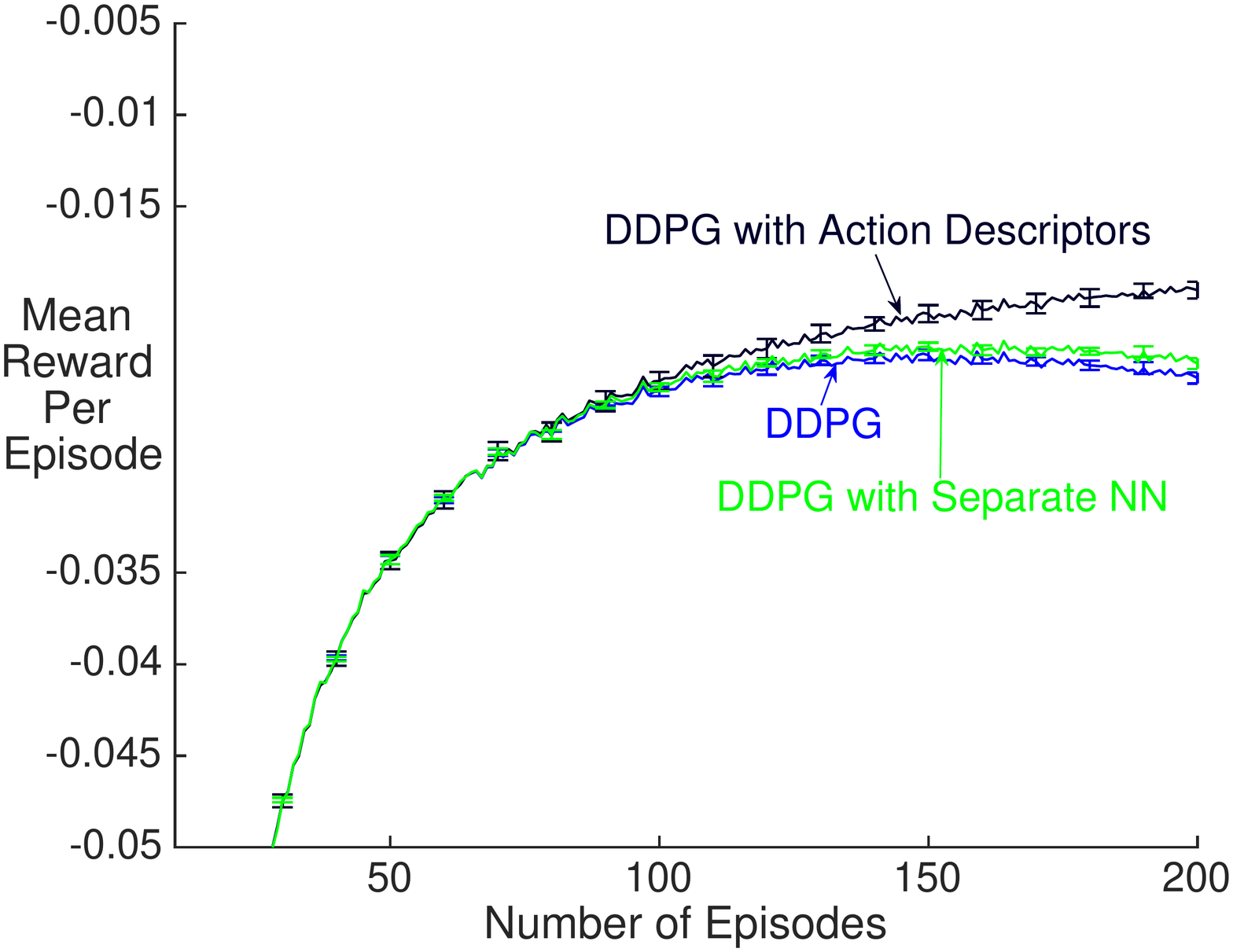}\label{fig:pdemodel_256d_l2}}
	\caption{
		Mean reward per episode vs. episodes on PDE-Model. The results are averaged over $30$ runs, with standard error displayed. 
	}\label{fig_pdemodel_l2}
	\vspace{-0.01cm}
\end{figure*}

The PDE Model domain is using the $2$D heat equation as the underlying transition dynamic (see \ifSupp Appendix~\ref{pde_details} \else the supplementary material \fi for details).
The infinite-dimensional state and action spaces are discretized to $\States \subset \Real^{d \times d}$ and $\Actions \subset [-1, 1]^{d \times d}$. 
Figure~\ref{fig_pdemodel_l2} shows the comparison between our DDPG with Action Descriptors and the other two competitors, as the number of state and action dimension increase: $d^2 \in \{36, 100, 256\}$. One can see that using action descriptors consistently outperforms the other algorithms. Both DDPG and the DDPG with separate NNs begin to decrease in performance after about 100 episodes, for higher-dimensional actions $d^2 = 100, 256$. DDPG with Action Descriptors, however, does not display this degradation for any action dimension, and continues improving with more training time.
Additionally, DDPG with separate NN shows slightly lower sample efficiency, as expected, though with higher action dimensions this effect is less obvious as all algorithms degrade and the regularity on this domain is not that strong. 

\subsection{Results on the Heat Invader Domain}
\begin{figure*}[ht!]
	\vspace{-0.5cm}
	\subfigure[25 dimensional action]{
		\includegraphics[width=\figwidththree]{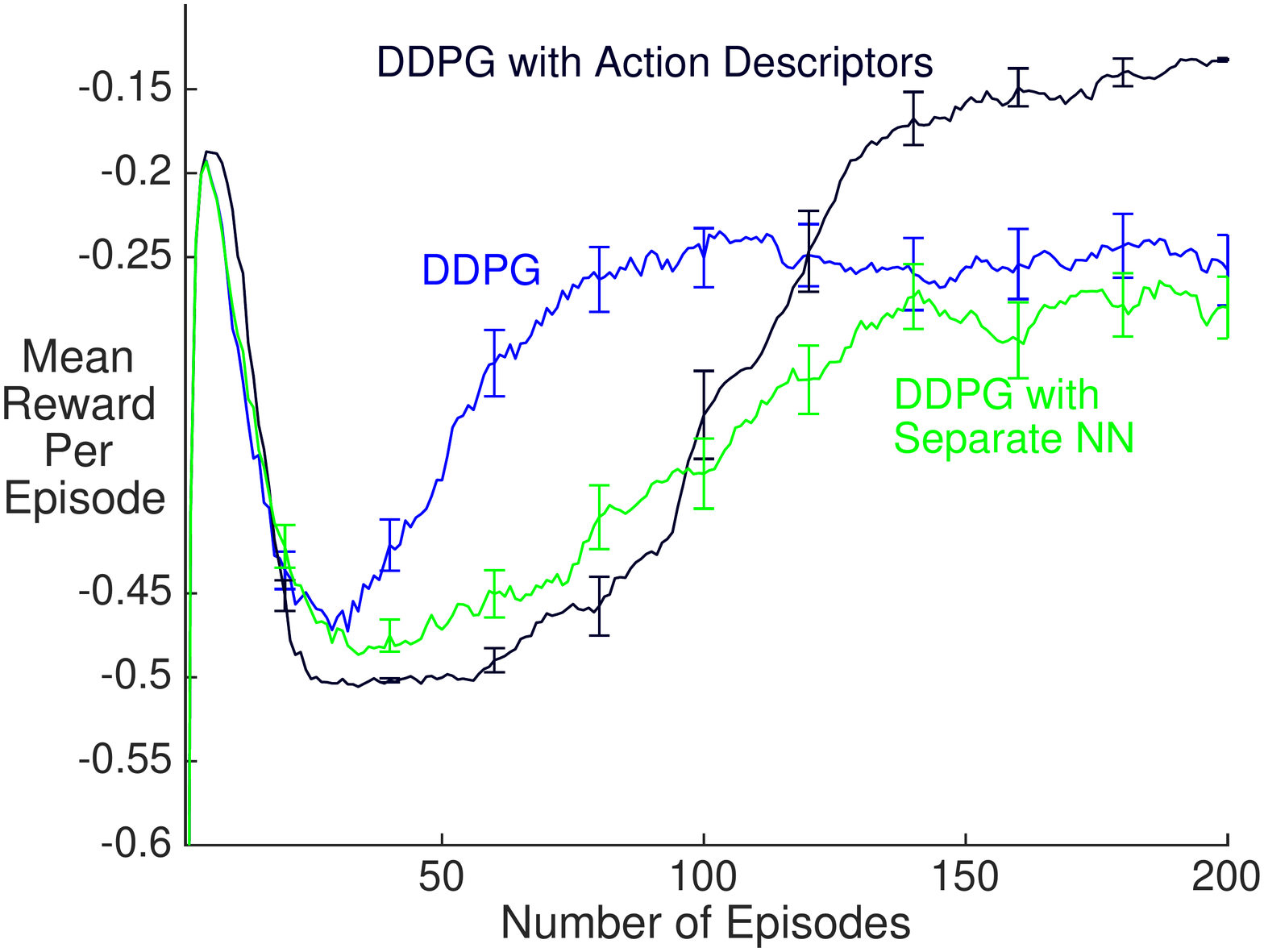}\label{fig:heat_25d_uniform}}
	\subfigure[50 dimensional action]{
		\includegraphics[width=\figwidththree]{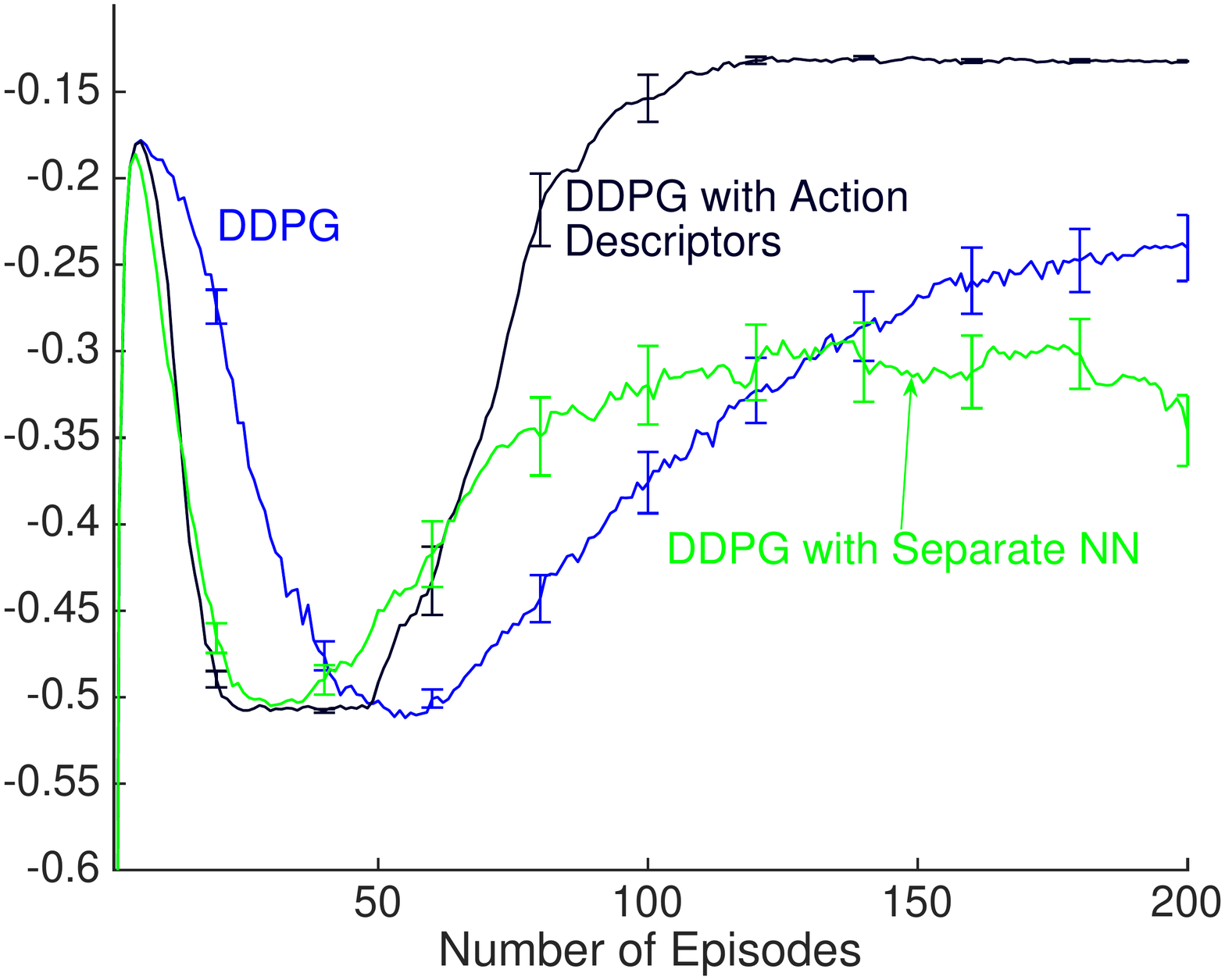}\label{fig:heat_50d_uniform}}
	\\
	\subfigure[100 dimensional action]{
		\includegraphics[width=\figwidththree]{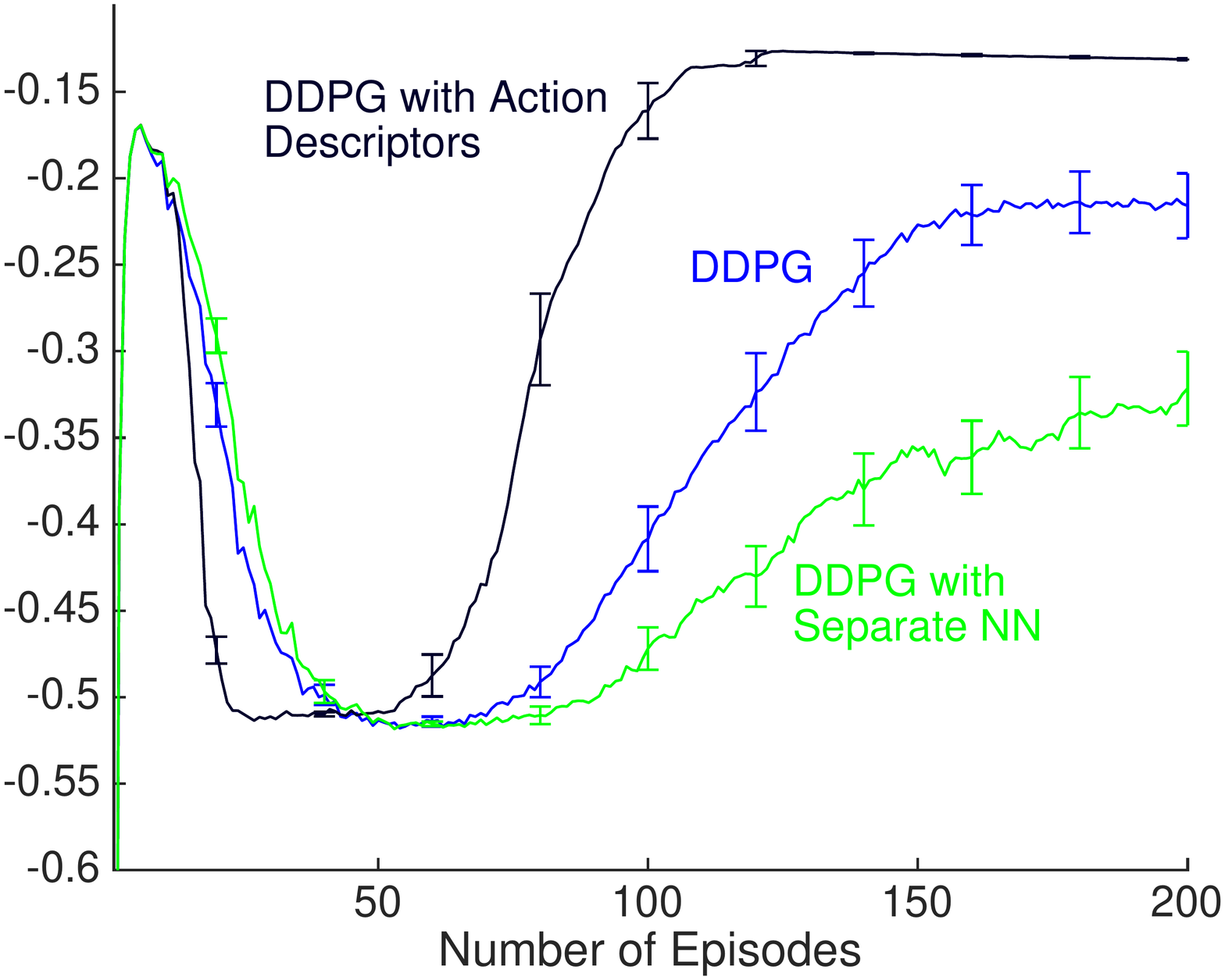}\label{fig:heat_100d_uniform}}
	\subfigure[200 dimensional action]{
		\includegraphics[width=\figwidththree]{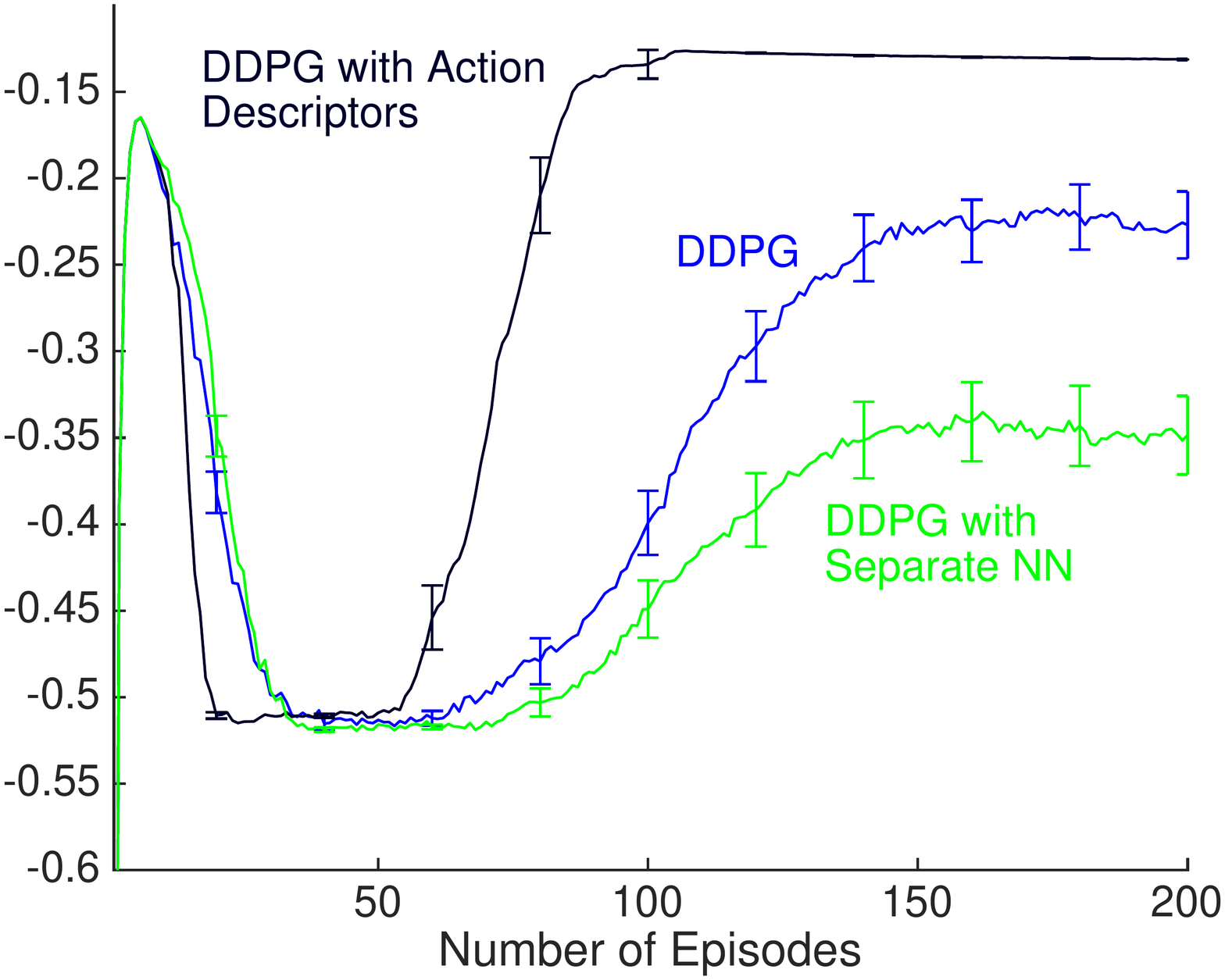}\label{fig:heat_200d_uniform}}
	\begin{minipage}{\figwidththree}
			\vspace{-8.5cm}	
		\caption{
			Mean reward per episode vs. episodes on the Heat Invader domain, with increasing action dimension. The results are averaged over $50$ runs. 
			 DDPG with action descriptors consistently outperforms other algorithms across dimensions, and DDPG shows higher sample efficiency than the DDPG with separate NN. We suspect the latter cannot converge to a good policy after the dimension increased to $50$ due to its inability to capture the regularities. The results suggest that $25$ air conditioners may not provide sufficiently fine resolution to control the temperature well in the room, as even the solution under DDPG with Action Descriptors has more variability. Note that the sharp increase in the reward early in the graphs is due to the relatively large noise in the action, triggering the airflow to be ON and making the temperature decrease faster. As the noise decreases after a few episodes, the fan turns off and the agent is improving its policy.
		}\label{fig:heat_uniform_compare}
	\end{minipage}
	\vspace{-0.5cm}
\end{figure*}


The underlying state transition dynamics of this domain is the PDE as described by Equation~\eqref{eq:conv_diff_pde}. 
In an ideal case, there are infinitely many air conditioners on the floor, and at each time step we can control the temperature for each air conditioner.
In simulation, performed using the finite volume solver FiPy~\citep{GuyerWheelerWarren2009},
the state and action spaces are discretized.
The state space is discretized to $\XX \subset \Real^{50 \times 50}$, giving a total of $2500$ measurement locations on the floor.  
Figure~\ref{fig:heat_uniform_compare} shows comparisons across different control dimensions  $\action_t \in [-0.5, 0]^k, k \in \{25, 50, 100, 200\}$. On this more difficult domain, it becomes clear that DDPG with separate NN has a lower sample efficiency than DDPG, and the former even shows divergence after the dimension reaches $50$. For $k = 100$ and $200$, DDPG with Action Descriptors learns a significantly better and more stable policy. Theoretically, the finer we can control the temperature, the higher reward we can potentially obtain. We suspect there is a tradeoff between how well the agent can learn the policy and how finely the temperature needs to be controlled. Once the dimension increases to $100$ or $200$, only DDPG with Action Descriptors is able to exploit this finer-grained control. 



\section{Conclusion}
\label{sec:HDARL-Conclusion}

We presented a general framework of using reinforcement learning for PDE control, which has infinite dimensional state and action spaces (or very high-dimensional in practice).
We proposed the concept of action descriptors to enable RL algorithms, such as deterministic policy gradient, to handle these problems. Theoretical evidence showed why our approach might have better sample efficiency compared to a conventional continuous-action RL approach. Our strategy enables the architecture to easily scale with increasing action dimension. We showed that the same neural network architecture, with the exception of the output layer, can obtain significantly better performance scaling as the dimension increases on two PDE domains. We believe that the proposed RL-based approach has the potential to be a powerful alternative to conventional control engineering methods to PDE control, as well as other RL/control problems with extremely high-dimensional action spaces with action regularities.

\section*{Acknowledgements}
We would like to thank the anonymous reviewers for their helpful feedback.


\ifSupp
\appendix
%

\section{Theoretical Analysis: Proofs and Additional Discussions}
\label{sec:HDARL-Appendix-Theory}

In this appendix we first review the notion of covering number (Appendix~\ref{sec:HDARL-Appendix-Covering}). Afterwards, we prove Proposition~\ref{prop:HDARL-CoveringNumber-LipschitzClasses}. The basic idea of the proof is to calculate the covering numbers of a policy space that has a Lipschitz continuity and that of another space without such property which still can provide an $\epsilon$-approximation to any object in the former policy space  (Appendix~\ref{sec:HDARL-Appendix-Covering-Proof}). Finally in Appendix~\ref{sec:HDARL-Appendix-NN-Lipschitz}, we relate the Lipschitz continuity of feedforward neural networks with their weights, thus showing that one can explicitly impose the Lipschitz continuity in DNN by controlling the norm of their weights.


\subsection{Covering Number}
\label{sec:HDARL-Appendix-Covering}

We briefly introduce the notion of covering number, and refer the reader to standard references for more discussions, e.g., Chapter 9 of~\citet{Gyorfi02}.
Consider a normed function space $\FF(\Omega)$ defined over the domain $\Omega$ with the norm denoted by $\norm{\cdot}$.
The set $S_\eps = \{f_1, \dots, f_{N_\eps} \}$ is the $\eps$-covering of $\FF$ w.r.t. the norm if for any $f \in \FF$, there exists an  $f' \in S_\eps$ such that $\norm{f - f'} \leq \eps$.
The covering number is the minimum $N_\eps$ such that there exists an $\eps$-cover $S_\eps$ for $\FF$. We use $\cN(\eps, \FF(\Omega))$ or simply $\cN(\eps)$ if $\FF(\Omega)$ is clear from the context, to denote it.

For simplicity of arguments in this paper, the covering number results use the supremum norm, which is defined as
\[
\norm{f}_\infty = \sup_{w \in \Omega} |f(w)|.
\]

\subsection{Proof for Proposition~\ref{prop:HDARL-CoveringNumber-LipschitzClasses}}
\label{sec:HDARL-Appendix-Covering-Proof}
\begin{proof}
	First we show that $\underline{\Pi}$ provides an $\eps$-approximation of $\Pi_L$.
	Pick any $\eps > 0$.
	For each $c_i$ ($i = 1, \dotsc, M_\eps$), pick a minimal $\eps/2$-covering set of $\Pi_L \rvert_{c_i}$, and call it $S_{\eps/2, c_i}$. Define the following function space:
	\begin{align*}
	\underline{\Pi}_{\eps/2} =
	\Big \{ &\pi_{\underline{\theta}}(x,z) = \sum_{i=1}^{M_\eps} \pi_{\theta_i}(x, c_i) \One{ z \in A_i}:  \\
	&\pi_{\theta_i} \in S_{\eps/2, c_i}, i=1, \dotsc, M_{\eps/2} \Big\}.
	\end{align*}
	This space is an $\eps/2$-covering of $\underline{\Pi}$. Moreover, it provides an $\eps$-covering of $\Pi_L$ too.
	To see this, consider any $\pi_\theta \in \Pi_L$.  
	For any fixed $(x, z) \in \XX \times \ZZ$, the location $z$ falls in one of the partitions $A_i$, so $\smallnorm{z - c_i} \leq \frac{\eps}{2L}$.
	By construction of $\underline{\Pi}_{\eps/2}$, there exists $\pi_{\underline{\theta}} \in \underline{\Pi}_{\eps/2}$ such that 
	$|\pi_\theta(x,c_i) - \pi_{\underline{\theta}}(x,c_i)| \leq \eps/2$.
	Therefore, we have 
	\begin{align*}
	\left | \pi_\theta (x,z) - \pi_{\underline{\theta}}(x,z) \right|
	\leq &
	\left |	\pi_\theta(x,z) - \pi_\theta(x,c_i)		\right| + {}
	\\
	&
	\left |	\pi_\theta(x,c_i) - \pi_{\underline{\theta}}(x,c_i)		\right| + {} 
	\\
	&
	\left| \pi_{\underline{\theta}}(x,c_i) - \pi_{\underline{\theta}}(x,z) \right|
	\\
	\leq &
	L \norm{z - c_i} + \frac{\eps}{2} + 0 \leq \eps.
	\end{align*}
	As we have $M_\eps$ cells, the number of members of the covering of $\pi_{\underline{\theta}}(x,c_i)$ is
	\begin{align}
	\label{eq:HDARL-CoveringArgument-Discretized-FA}
	\left[ \cN (\eps/2) \right]^{M_\eps}.
	\end{align}
	Substituting the value of $M_\eps$ leads to the desired result for $\cN(\eps,\underline{\Pi})$.
	
	To derive the covering number for $\Pi_L$, let $\eps > 0$ and define the following function space:
	\begin{align*}
	\label{eq:HDARL-Proof-tildePi}
	\tilde{\Pi}_\eps = \{\tilde{\pi}(x,z) = \sum_{i=1}^{M_\eps} 0.5 \eps \left \lfloor \frac{ \pi(x,c_i)}{0.5\eps} \right \rfloor \One{z \in A_i}, \pi \in \Pi_L \} 
	\end{align*} 
	We shortly show that $\tilde{\Pi}_\eps$ is an $\eps$-covering of $\Pi_L$.
	Also notice that $\tilde{\pi}$ only takes discrete values with resolution of $\eps/2$.

	Consider any $\pi_\theta \in \Pi_L$.
	As in the previous case, for any fixed $(x, z) \in \XX \times \ZZ$, the location $z$ falls in one of the partitions $A_i$. 
	Because of the construction of $\tilde{\Pi}_\eps$, there exists a $\tilde{\pi}$ that is $\eps/2$-close to $\pi(x,c_i)$. Using this property and the Lipschitzness of $\pi_\theta \in \Pi_L$, we have
	\begin{align}
	\nonumber
	| \pi_\theta (x,z) - \tilde{\pi}(x,z) | \leq & 
	| \pi_\theta(x,z) - \pi_\theta(x,c_i) | +
	\\
	\nonumber
	&
	|\pi_\theta (x,c_i) - \tilde{\pi}(x,c_i) | + \\
	\nonumber&
	|\tilde{\pi}(x,c_i) - \tilde{\pi}(x,z) |
	\\ 
	\nonumber
	\leq &
	L \norm{z - c_i} + \\
	&
	\nonumber
	\left| 0.5\eps \left \lfloor \frac{ \pi_\theta(x,c_i)}{0.5\eps} \right \rfloor - \pi_\theta (x,c_i) \right| + 0 \\
	& \leq \eps/2 + \eps/2 = \eps
	\end{align}
	
	So $\tilde{\Pi}_\eps$ provides $\Pi_L$ with an $\eps$-cover. It remains to count the number of elements of $\tilde{\Pi}_\eps$.
	
	We choose an arbitrary centre $c_1$. We let the function $\tilde{\pi}(x,c_1)$ to be one of possible $\cN(\eps)$ members of the minimal covering of $\Pi_L \rvert_{c_1}$. Notice that for any $c_i$ and $c_j$ that are neighbour (so they have distance less than $\eps / L$), using the Lipschitzness of $\pi$ and the definition of $\tilde{\pi}$, we have
	\begin{align*}
	\left | \tilde{\pi}(x,c_i) - \tilde{\pi}(x,c_j) \right |
	\leq &
	\left | \tilde{\pi}(x,c_i) - \pi(x,c_j) \right | + \\
	&
	\left | \pi(x,c_i) - \pi(x,c_j) \right | + \\
	&
	\left | \pi(x,c_j) - \tilde{\pi}(x,c_j) \right |
	\\
	\leq & 2 \eps.
	\end{align*}
	
	%
	%

	Consider two neighbour centres $c_i$ and $c_j \neq c_i$. Since $\tilde{\pi}$ only takes discrete values (with the resolution of $\eps/2$), the value of $\tilde{\pi}(x,c_j)$ can only be one of $9$ possible values, i.e., $\tilde{\pi}(x,c_i) - 4\eps/2, \tilde{\pi}(x,c_i) - 3\eps/2, \dotsc, \tilde{\pi}(x,c_i) + 4 \eps/2$.
	
	Choose $c_i = c_1$. One of its neighbour, let us call it $c_j = c_2$, can take at most $9$ different values. Therefore, the total number of function $\tilde{\pi}$ defined over $\XX \times \{ c_1, c_2 \}$ is $9 \cN(\eps)$.
	We continue this argument with an arbitrary sweep through neighbourhood structure of the partition. As there are at most $M_\eps$ points, the number of possible functions $\tilde{\pi}_\eps$ is
	\begin{align}
	\label{eq:HDARL-CoveringArgument-LipschitzSpace}
	\cN(\eps) \times 9^{M_\eps}.
	\end{align}
	Replacing the value of $M_\eps$ leads to the desired result.
\end{proof}
We would like to acknowledge that the argument for counting the members of $\tilde{\Pi}_\eps$ is borrowed from a similar argument in Lemma~2.3 of~\citet{vandeGeer00}.

\subsection{Lipschitzness of Feedforward Neural Networks}
\label{sec:HDARL-Appendix-NN-Lipschitz}
The function space $\Pi_L$ in Section~\ref{sec:HDARL-Theory} is $L$-Lipschitz in the action location $z$. 
To show how one might impose this condition in practice, we focus on NN and relate the Lipschitz constant of a feedforward NN to its weights. Consequently, by controlling the weights, we might control the Lipschitz constant of the NN.
Note that the Lipschitz properties of NN has been studied before, either as a standalone result (e.g.,~\citealt{BalanSinghZou2017,AsadiMisraLittman2018}) or as a part of proof for another result (e.g., \citealt{Barron1994}).

A single layer of feedforward NN with $\text{ReLU}$ or $\tanh$ nonlinearity is Lipschitz.
To see this, consider an $\ell_p$-norm and its induced matrix norm (for any $1 \leq p \leq \infty$).
As $\text{ReLU}$ or $\tanh$ are both $1$-Lipschitz coordinate-wise, we have
\begin{align*}
	\norm{ \text{ReLU}(W x_1) - \text{ReLU}(W x_2) }_p
	& \leq
	\norm{
		W (x_1 - x_2)
	}_p
	\\
	&
	\leq
	\norm{W}_{p} \norm{x_1 - x_2}_p.
\end{align*}
Therefore, its Lipschitz constant is $L = \norm{W}_p$.
As the composition of Lipschitz functions with Lipschitz constants $L_1$, $L_2$, etc. w.r.t. the same norm is also Lipschitz with the Lipschitz constant $L = L_1 L_2 \cdots$, one may obtain the overall Lipschitz constant of a multi-layer feedforward NN.
So by controlling the norm of the weights, either by regularization or weight clipping, one can control the overall Lipschitz constant of the network.




\section{Detail of Experiments}

In this section, we first introduce concrete neural network and parameter settings followed by the details of how we optimize each algorithm and how we plot the learning curve. Afterwards, we provide details about the PDE Model and the Heat Invader domains. Additional experimental results are given for completeness. 
\todo{Maybe a short description of what we present here. -AMF}

\subsection{Experimental Settings}

\textbf{Neural network architecture.} We follow the same neural network architecture as introduced in the original DDPG paper \citep{lillicrap2016}. We use TensorFlow~\citep{tensorflow2015-whitepaper} for our implementations.
The DNN has $3$ convolutional layers without pooling, and each layer has 32 filters. The kernel sizes are $4\times4$ for the first two layers and $3\times3$ for the third one. Batch normalization~\citep{sergey2015batchnorm} is used in the convolutional layers. The convolutional layers are followed by two fully connected layers and each has $200$ ReLU units. In the actor network, the activation function in the output layer is \emph{sigmoid} for the Heat Invader domain and \emph{tanh} for the PDE Model domain. The critic network uses the same convolutional layer setting, but its activation function in the output layer is linear. The $L_2$ weight decay with parameter $0.001$ is used after the convolutional layers in the critic network. The final layer parameters of both  actor and critic networks were initialized from a uniform distribution $[-0.0003, 0.0003]$ and all other parameters are initialized using Xavier initialization~\citep{xavier2010deepnn}.

As for the DDPG with separate NN, if the action is in $\Real^k$, we have $k$ neural networks with the same architecture as explained above, except that all of them share the same convolutional layers.

For DDPG with action descriptors, we use exactly the same architecture as DDPG, except that the output layer has only one output unit. The \emph{action descriptor} comes into the neural network immediately after the convolutional layers by concatenating. 

\textbf{Parameter setting.} Across all experiments, we used experience replay with the buffer size of $20,000$ and the batch size of $16$. Each episode has $40$ steps. The discount factor is $0.99$. The exploration noise is from the Gaussian distribution $\mathcal{N}(u_t, \frac{1.0}{\text{episodes}})$. For each choice of action dimension, we sweep over parameters as follows. Actor learning rate is selected from $\alpha \in \{0.000001, 0.000005, 0.00001, 0.0001\}$, critic learning rate is from $\alpha \times \{1.0, 5.0, 10.0, 20.0, 40.0\}$. The intuition behind this selection is that the critic learning rate should be larger, and in fact our experimental results are consistent with this intuition.

\textbf{Performance measure}. We take a similar approach as in the previous work~\citep{FarahmandNabiNikovski2017}, computing the mean reward per step for each episode $i$ averaged over $N$ runs. Hence on our learning curve, given episode index $i$, the corresponding $y$-axis value is computed as $$R^{(i)} = \frac{1}{N}\sum_{n = 1}^{N} \frac{1}{T}\sum_{t= 1}^{T} r^{(n, i)}_t,$$ where $T = 40$ across all experiments. To pick up the best parameter, we compute $\text{Evaluate} = \sum_{i = m}^{n} R^{(i)}$. For the PDE Model, we use $m = 180$ and $n = 200$, whereas for the Heat Invader, we use $m = 150$ and $n = 200$. We choose these ranges to avoid a ``divergent" learning curve
which may happen when an algorithm picks a large learning rate and hence converges quickly at the beginning but later diverges. 

\subsection{Additional Details on PDE Model}\label{pde_details}

The PDE Model domain is a simple model that allows easy experimentation. It is defined based on the controlled $2$D heat equation
\begin{equation*}
	\frac{\partial h(z, t)}{\partial t} = \alpha (\nabla^2 h + a),
\end{equation*} 
\todo{If $a$ is not multiplied by $\alpha$ here, why do we have it inside the parentheses in the next finite-difference equation? -AMF Updated, it is more reasonable to put action inside}
where $h$ is the heat, a function of location $z = (x, y) \in \Real^2$ and time $t$, $a$ is a function-valued action (or control signal), and 
$\alpha$ is a constant that depends on the physical properties of the environment (thermal conductivity, etc.) and is called thermal diffusivity. The action $a$ is a scalar field that we can control in order to modify the behaviour of the dynamical system.
We let $\alpha= 1$ for simplicity. 
\todo{So we need to add the action here too. I changed it. Please verify that this is indeed the case.}

We use the finite difference time domain (FDTD) method to solve this PDE, which we now describe.
We discretize the 2D spatial domain into $d \times d$ elements with $d$ being an integer value. This can be represented by a matrix, in which each entry indicates the heat value at a particular location.
%
We can implement the $d \times d$-dimensional state transition (still in continuous time) as
\begin{align*}
& \frac{\partial h(z, t)}{\partial t} =\\
& \alpha \Big(\!\frac{h(x\!+\!\delta, y, t)\!+\!h(x\!-\!\delta, y, t)\!+\!h(x, y\!-\!\delta, t)\! + \!h(x, y\! +\! \delta, t)}{c} \\
& - \frac{4 h(x,y,t)}{c} + \text{action}(x,y,t) \Big),
\end{align*} 
where $c$ is a constant to scale the change \todo{Is $c$ equal to $\delta^2$? -AMF It is used for numerical approximation, i want to differentiate this continuous formulae with the below discretized version. $c$ has the effects from both below $\delta_t$ and $\delta_s$} and the $\text{action}(x,y,t)$ is the control command that is under our control at location $(x,y)$.
For each $\state \in \States$, let  $\state^{(i,j)}$ denotes the element in the $i$-th row and $j$-th column of the matrix $\state$. 
To turn this continuous-time finite difference approximation to discrete-time finite difference approximation (known as finite-difference time domain (FDTD) method), let us denote $\state_t$ and $\action_t$ as the state and action at time $t$.
The state dynamics are described by the following equations, 
\begin{align*}
\!\!&\Delta x_t \!=\! \frac{ \! \left(\! \stateind{t}{i-1}{j} \!+\! \stateind{t}{i+1}{j}  \!+\! \stateind{t}{i}{j-1} \! +\! \stateind{t}{i}{j+1} \!  - \!4\stateind{t}{i-1}{j-1} \right)}{\delta_s} \! \\
& \qquad + \action^{(i-1, j-1)}_t \\
&\stateind{t+1}{i-1}{j-1} \gets \stateind{t}{i-1}{j-1} + \delta_t \Delta x_t,
\end{align*}
with the temporal discretization $\delta_t\!=\!0.001$ and spatial discretization $\delta_s\!=\!0.1$ for the finite difference method. \todo{What is the difference between $\delta_s$ and $c$ above? And what about $\alpha$ and $\delta_t$? I think this part can be cleared a bit. -AMF $\alpha$ is the thermal diffusivity depending on physical material}

For boundary values, where $i-1$ or $i+1$ become smaller than 0 or larger than $d$, we set them to zero, i.e., $\stateind{t}{i}{j} = 0, \forall i, j \notin  \{1, ..., d\}$. The reward function is defined as%
\begin{equation*}
r(\state_t, \action_t, \state_{t+1}) = -\frac{||\state_{t+1}||_2}{d} - \frac{||\action_t||_2}{d}.
\end{equation*}

One can intuitively understand this equation as following. Since the heat always move from the high heat region to low heat region, for each entry in the matrix (i.e. a location on the $2$-D room) we can take the sum of its neighbors as the heat value coming from nearby region. Then if that sum is larger than the entry's current value, we should expect the entry's heat increases.
At the beginning of each episode, each entries' values are independently and uniformly initialized from $[0,1]^{d\times d}$. Note that at each step, the update to the value in each entry happens simultaneously. The boundary values are set as $0$ across all steps. Each episode is restricted to $40$ steps and at each step, an action is repeatedly executed $100$ times. This design is based on the intuition that the agent may not react quickly enough to change action every $0.001$ time unit. 

Our design of reward function is mainly based on the intuition that we want to keep the temperature low and action cost small. We believe other reasonable reward choices can also work. 

We choose action descriptors simply by uniformly choosing from the space $[-0.5, 0.5]^2$. There is no particular reason for our choice, we believe other range such as $[-1, 1]$ can also work, as long as the descriptors are topologically distributed the same as the air conditioners. 

\subsection{Additional Details on Heat Invader}
\label{sec:heat_details}

\begin{figure}[t]
	\includegraphics[width=\figwidthtwo]{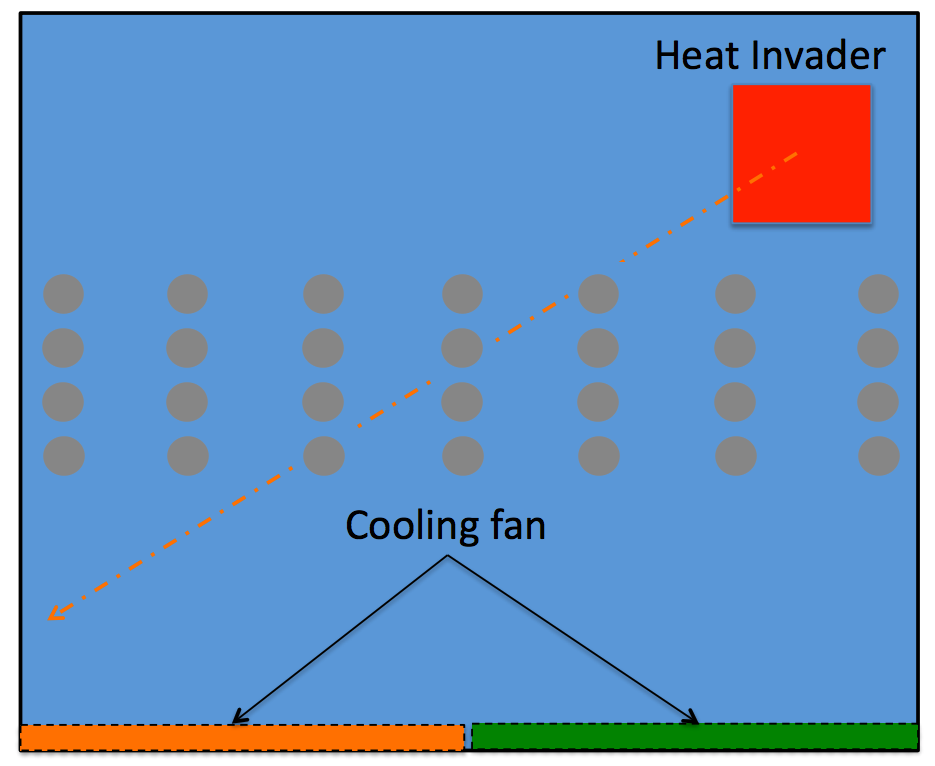}
	\raggedright
	\caption{
		\raggedright
		Heat invader domain. Image is modified from the paper~\citep{FarahmandNabiGroverNikovski2016}. The solid circle refers to the elements of the air conditioner. The actual number of air conditioners is different from what the figure shows.
	}\label{heatinvaderdomain}
\end{figure}

We include Figure~\ref{heatinvaderdomain} to describe the domain: Heat Invader. The action is discretized to four rows, lying in the middle of the room and each row has $50$ air conditioners. The output action is repeated to $200$ executable action dimensions if the output dimension is less than $200$. Similar to~\citet{FarahmandNabiNikovski2017}, we still design two fans symmetrically located on the left and right wall of the room. A fan will be trigged if the sum of absolute values of actions located on that corresponding half area of the room exceeds some threshold, which we set as $25$. The reward function
\begin{align}
r(\state_t, \action_t, &\state_{t+1}) =  - \text{cost}(\action_t)\\
& - \int_{z \in \ZZ}^{} I ({|T(z, t+1)| > T^\ast(z, t+1)}) \mathrm{d}z \nonumber
\end{align}
is designed as following. $\ZZ$ is discretized to $50 \times 50$ on the floor and hence there are totally $2500$ temperature measurements. We set $T^\ast(z, t+1) = 0.501, \forall z \in \ZZ$. We further scaled the integration (summation over locations) by $2500$. The cost from action is defined as $\text{cost}(a_t) = \frac{||a_t||}{d}, a_t \in \Real^d$. The initial position of the heat invader is randomly sampled from $(i, j) \in [45, 50]^2, i, j \in \mathbb{Z}$. There are two types of airflow to choose, one is \emph{uniform} and another one is \emph{whirl}. We also include results by using \emph{whirl} airflow as showed in Figure~\ref{fig:heat_whirl_compare}. Note that there is a common learning pattern when using either uniform or whirl airflow. At the beginning, the exploration noise is large and hence the fan is triggered frequently, allowing the temperature to be lowered faster. As the noise decreasing, the penalty coming from the ``uncomfortable temperature'' shows effect and hence the performance drops. Afterwards while the agent improves the policy, the performance starts to increases again. 

We design the set of action descriptors as follows. When the action dimension $k \le 50$, we think the air conditioners are located on a single line in the middle of the room, hence we pick the descriptors uniformly from the segment $[-1, 1]$. When the action dimension $k = 100$, we think there are two lines on the middle of the room and hence we design the descriptors as two dimensional vectors $(x, y) \in [-1, 1]^2$, where $x$ takes two values $-1, 1$ while $y$ takes $50$ values uniformly along $[-1, 1]$. Similar approach applied to the case $k = 200$. Note that this domain has a fixed executable action dimension $a_t \in \Real^{200}$, hence when the output action from the agent has dimension less than $200$ (i.e., $1, 25, 50, 100$), the \emph{adapter} $I(\mathcal{C}, u_t)$ would map the output action to $200$ dimensions by repeating.

As a sanity check, we also conducted experiments when we use only $1$ dimensional action as showed in figure \ref{fig:heat_whirl_compare}. Neither of the two algorithms can learn a good policy for $k=1$, because there is no sufficiently fine-grained control. In this case, all three algorithms are almost identical, except that DDPG with Action Descriptors has two additional input units, as the descriptor is in $\Real^2$. 

Due to computational resource restriction, to generate the figure using the uniform airflow, we use $10$ runs to find best parameter settings and then do $50$ runs for that best setting.

\begin{figure*}[htp!]
	\subfigure[25 dimensional action]{
		\includegraphics[width=\figwidthtwo]{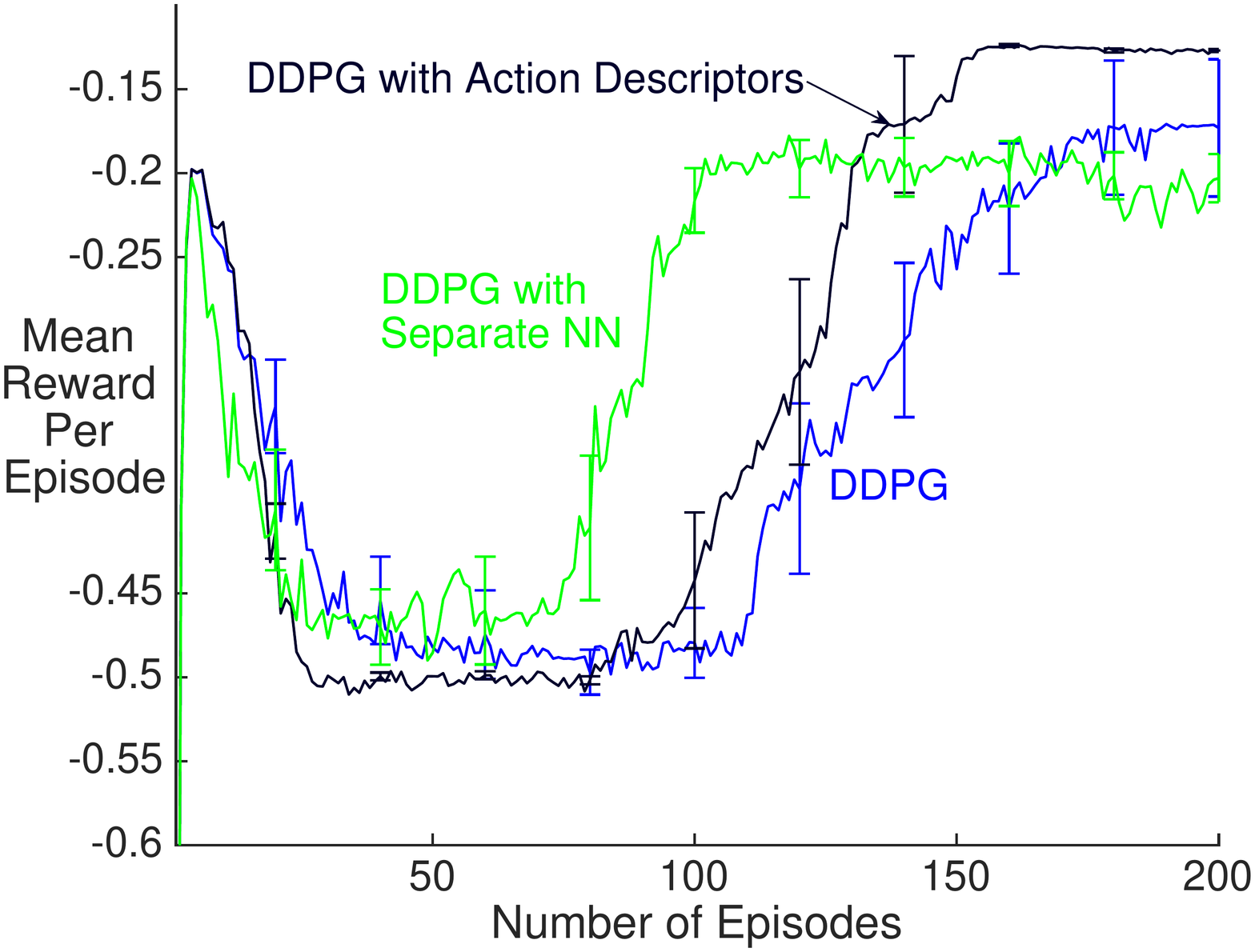}\label{fig:heat_25d_whirl}}
	\subfigure[50 dimensional action]{
		\includegraphics[width=\figwidthtwo]{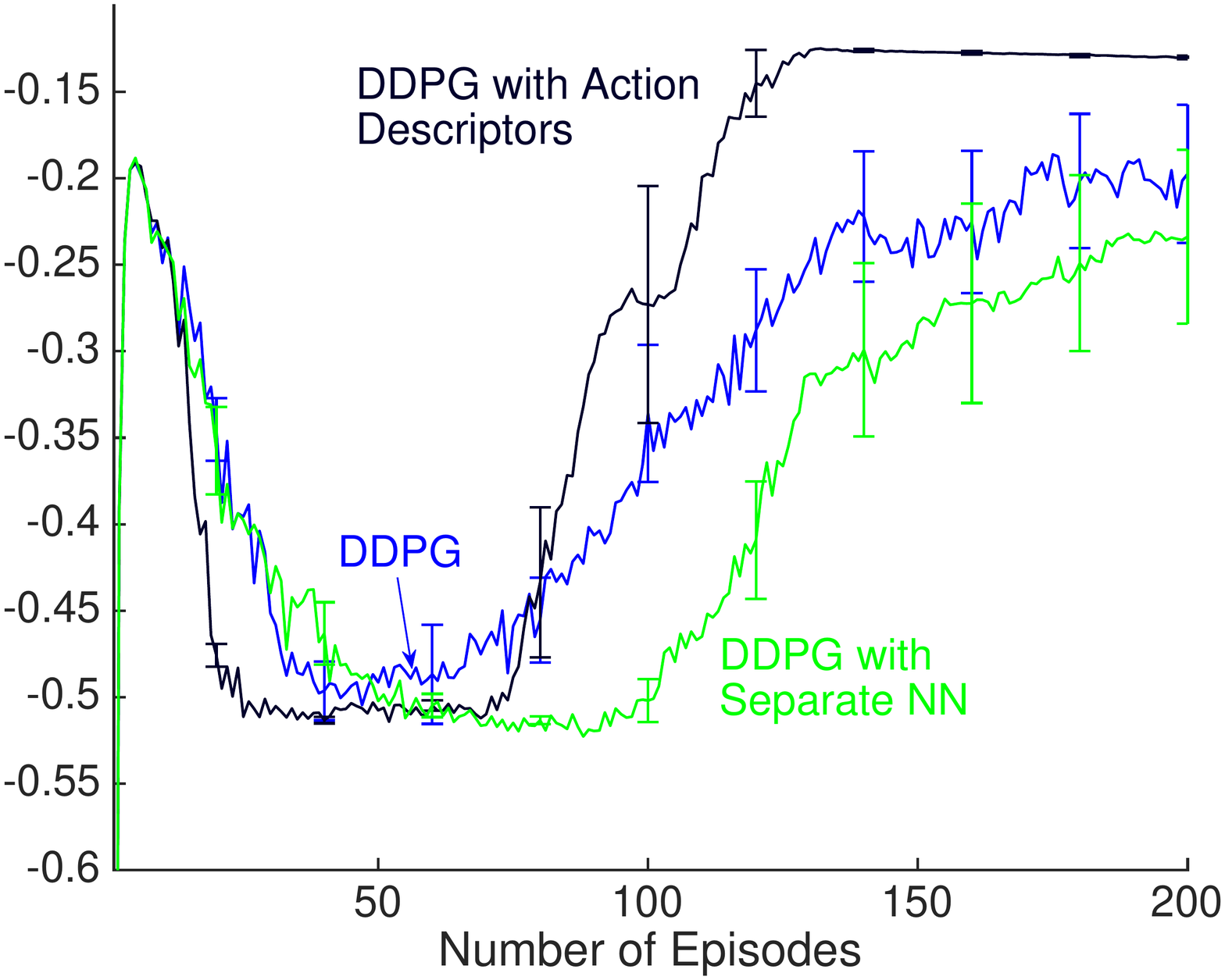}\label{fig:heat_50d_whirl}}
	\subfigure[100 dimensional action]{
		\includegraphics[width=\figwidthtwo]{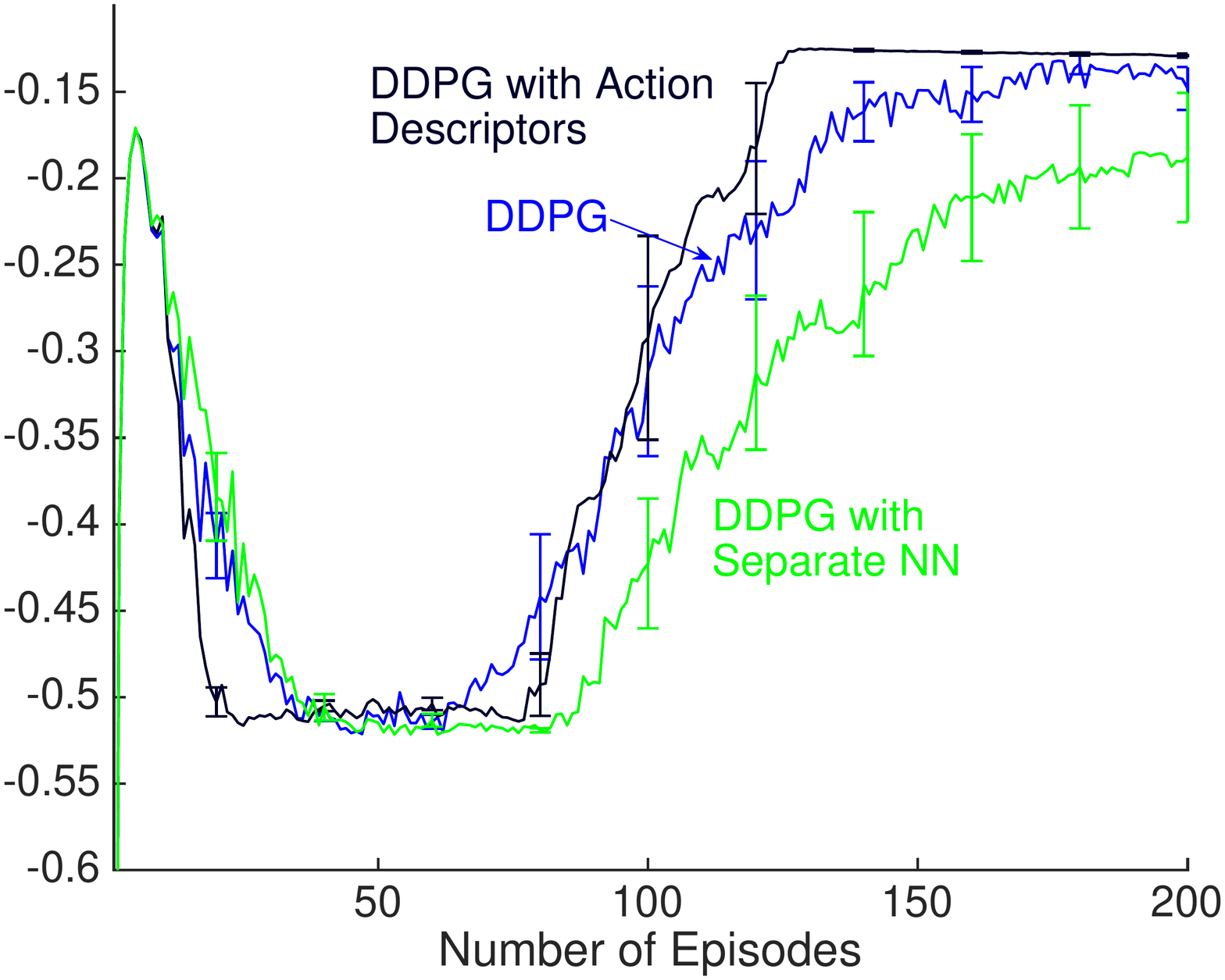}\label{fig:heat_100d_whirl}}
	\subfigure[200 dimensional action]{
		\includegraphics[width=\figwidthtwo]{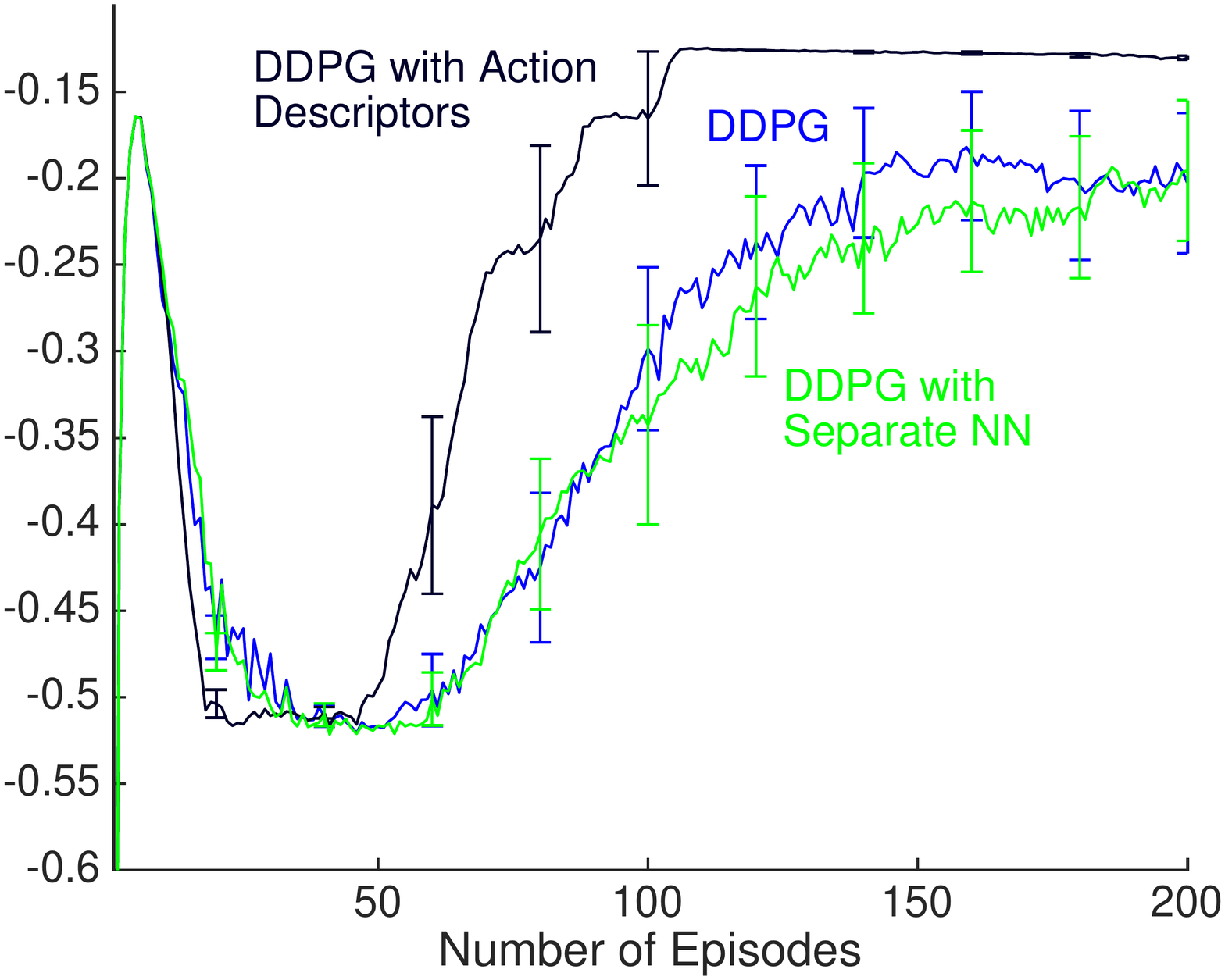}\label{fig:heat_200d_whirl}}
	\subfigure[1 dimensional action, uniform]{
		\includegraphics[width=\figwidthtwo]{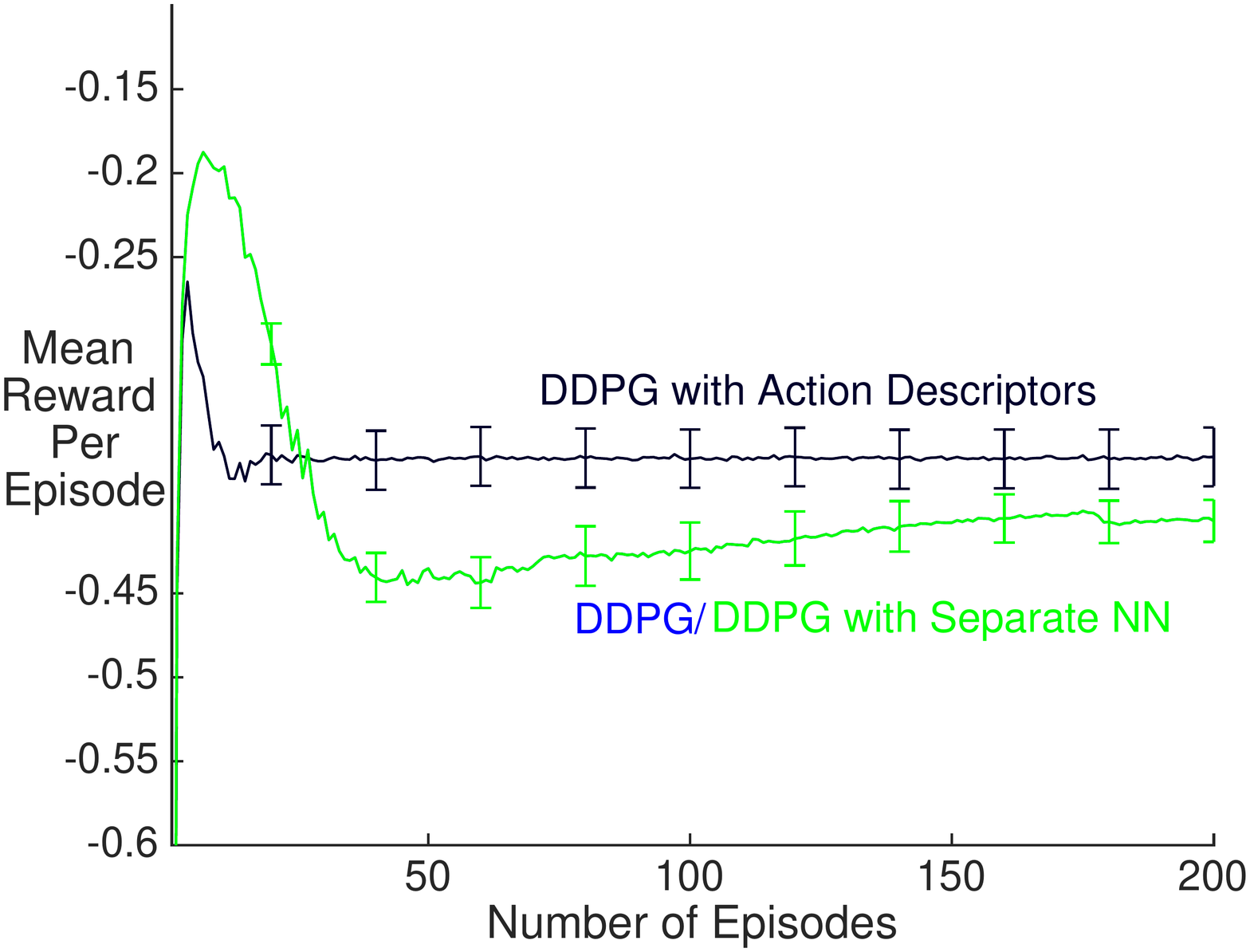}\label{fig:heat_1d_uniform}}
	\subfigure[1 dimensional action, whirl]{
		\includegraphics[width=\figwidthtwo]{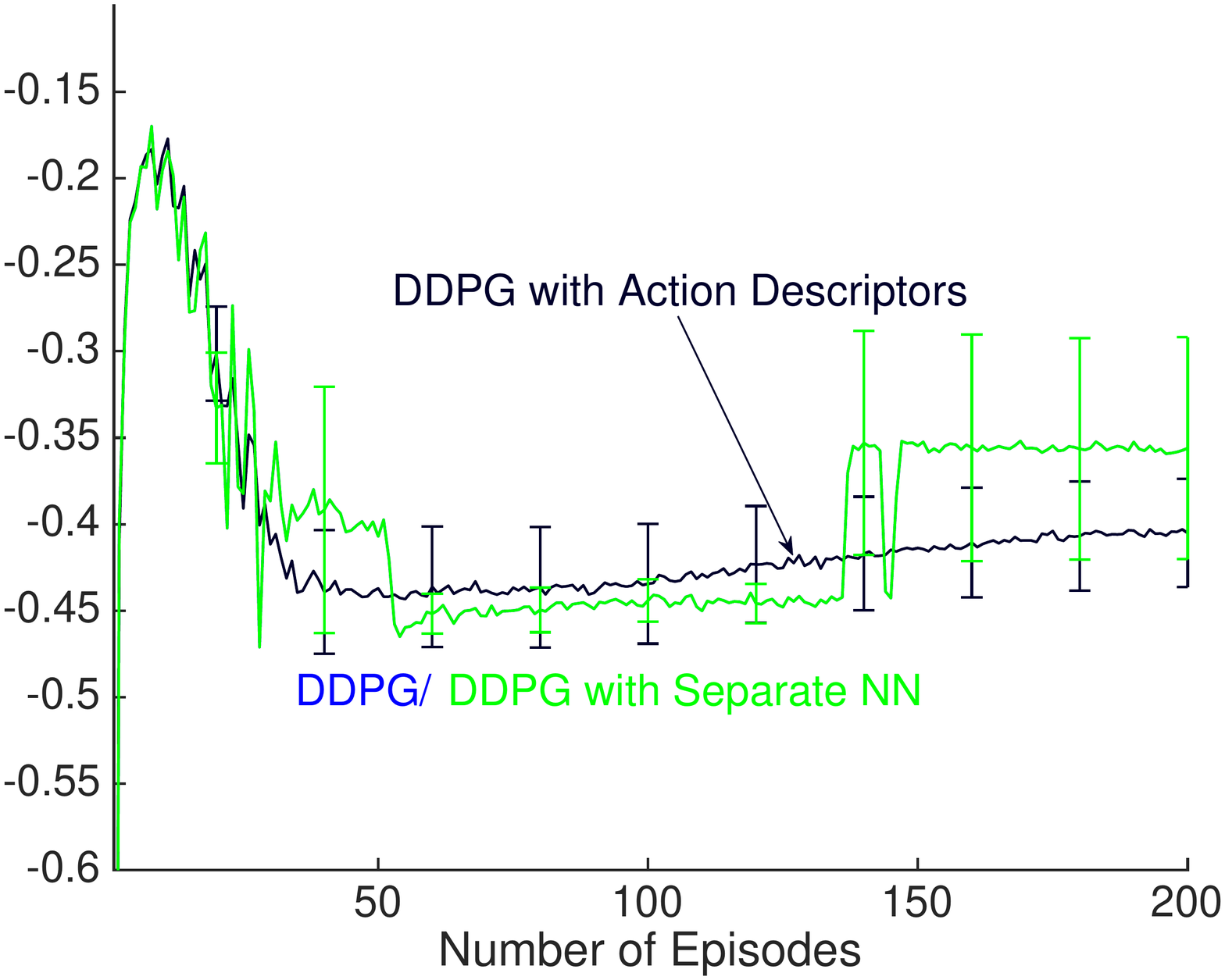}\label{fig:heat_1d_whirl}}
		\caption{
			Results of mean reward per episode vs. episodes on the Heat Invader domain, with an increasing number of action dimensions. The results are averaged over $10$ runs except Figure (e). 
			This figure is generated by using \emph{whirl} air flow unless otherwise specified. The results basically match with what we see in the experimental section, which was generated using \emph{uniform} airflow. Figure (e) and (f) show the comparison when using $1$d action dimension as a sanity check. 
		}\label{fig:heat_whirl_compare}
\end{figure*}
\fi

\clearpage
\newpage
{
\small
\bibliography{MyBib,paper} 
\bibliographystyle{icml2018}
}

\end{document}